\documentclass[a4paper,twoside]{article}

\usepackage{epsfig}
\usepackage{subcaption}
\usepackage{calc}
\usepackage{amssymb}
\usepackage{amstext}
\usepackage{amsmath}
\usepackage{amsthm}
\usepackage{multicol}
\usepackage{pslatex}
\usepackage{apalike}
\usepackage[bottom]{footmisc}
\usepackage{dblfloatfix}
\usepackage[table]{xcolor}
\usepackage{color, colortbl}
\usepackage{adjustbox}

\usepackage{ScitePRESS}     

\usepackage{amsmath,amsfonts,bm}

\def\eqref#1{(\ref{#1})}

\def\1{\bm{1}}

\def\vx{{\bm{x}}}
\def\vy{{\bm{y}}}

\def\mJ{{\bm{J}}}

\def\mX{{\bm{X}}}

\DeclareMathAlphabet{\mathsfit}{\encodingdefault}{\sfdefault}{m}{sl}
\SetMathAlphabet{\mathsfit}{bold}{\encodingdefault}{\sfdefault}{bx}{n}

\newcommand{\corrected}[1]{{\color{red} #1}}

\begin{document}

\title{Unfolding Local Growth Rate Estimates\\ for (Almost) Perfect Adversarial Detection}


\author{\authorname{Peter Lorenz\sup{1,2}, Margret Keuper\sup{3,4} and Janis Keuper\sup{1,5}}
\affiliation{\sup{1} ITWM Fraunhofer, Kaiserslautern, Germany}
\affiliation{\sup{2} Heidelberg University, Germany}
\affiliation{\sup{3} University of Siegen, Germany}
\affiliation{\sup{4} Max Planck Institute for Informatics, Saarland Informatics Campus, Saarbrücken, Germany}
\affiliation{\sup{5} IMLA, Offenburg University, Germany}
\email{Correspondence to peter.lorenz@itwm.fhg.de}
}


\abstract{
Convolutional neural networks (CNN) define the state-of-the-art solution on many perceptual tasks. However, current CNN approaches largely remain vulnerable against adversarial perturbations of the input that have been crafted specifically to fool the system while being quasi-imperceptible to the human eye. In recent years, various approaches have been proposed to defend CNNs against such attacks, for example by model hardening or by adding explicit defense mechanisms. Thereby, a small ``detector'' is included in the network and trained on the binary classification task of distinguishing genuine data from data containing adversarial perturbations. In this work, we propose a simple and light-weight detector, which leverages recent findings on the relation between networks' local intrinsic dimensionality (LID) and adversarial attacks. Based on a re-interpretation of the LID measure and several simple adaptations, we surpass the state-of-the-art on adversarial detection by a significant margin and reach almost perfect results in terms of F1-score for several networks and datasets. \\
Sources available at: \url{https://github.com/adverML/multiLID} 
}

\onecolumn \maketitle \normalsize \setcounter{footnote}{0} \vfill

\noindent\fcolorbox{red}{white}{
\begin{minipage}{0.99\columnwidth}
Erratum: Reported experimental results in all tables and figures have been corrected after finding and fixing errors in the implementation of multiLID and LID features. All changes are marked in red.
\end{minipage}
}

\section{{Introduction}} \label{introduction}

Deep Neural Networks (DNNs) are highly expressive models that have achieved state-of-the-art performance on a wide range of complex problems, such as in image classification. However, studies have found that DNNs can easily be compromised by adversarial examples \cite{fgsm,pgd,fabtattack,autoattack}. Applying these intentional perturbations to network inputs, chances of potential attackers fooling target networks into making incorrect predictions at test time are very high  \cite{carliniefficient}. Hence, this undesirable property of deep networks has become a major security concern in real-world applications of DNNs, such as self-driving cars and identity recognition \cite{stop_sign,Sharif_2019}.\\

\noindent Recent research on adversarial countermeasures can be grouped into two main approach angles: adversarial training and adversarial detection. While the first group of methods aims to "harden" the robustness of networks by augmenting the training data with adversarial examples, the later group tries to detect and reject malignant inputs.\\

\noindent In this paper, we restrict our investigation to the detection of adversarial images exposed to convolutional neural networks (CNN). We introduce a novel white-box detector, showing a close-to-perfect detection performance on widely used benchmark settings. Our method is built on the notion that adversarial samples are forming distinct sub-spaces, not only in the input domain but most dominantly in the feature spaces of neural networks \cite{szegedy13}. Hence, several prior works have attempted to find quantitative measures for the characterization and identification of such adversarial regions. We investigate the properties of the commonly used {\it local intrinsic dimensionality (LID)} and show that a robust identification of adversarial sub-spaces requires (i) an unfolded local representation and (ii) a non-linear separation of these manifolds. We utilize these insights to formulate our novel {\it multiLID} descriptor. Extensive experimental evaluations of the proposed approach show that {\it multiLID} allows reliable identification of adversarial samples generated by state-of-the-art attacks on CNNs. 
In summary, our contributions are:
\begin{itemize}
    \item an analysis of the widely used LID detector.
    \item novel re-formulation of an unfolded, non-linear {\it multiLID} descriptor which allows a close to perfect detection of adversarial input images in CNN architectures.  
    \item in-depth evaluation of our approach on common benchmark architectures and datasets, showing the superior performance of the proposed method.
\end{itemize}

\section{Related Work} \label{relatedwork}
In the following, we first briefly review the related work on adversarial attacks and provide details on the established attack approaches that we base our evaluation on. Then, we summarize approaches to network hardening by adversarial training. Last, we revise the literature on adversarial detection. 
\subsection{Adversarial Attacks} \label{sec:adversarial_attacks} Convolutional neural networks are known to be susceptible to adversarial attacks, i.e. (usually small) perturbation of the input images that are optimized to flip the network's decision. Several such attacks have been proposed in the past and we base our experimental evaluation on the following subset of the most widely used attacks. \\

     \noindent\textbf{\acf{fgsm}\,} \cite{fgsm} uses the gradients of a given model to create adversarial examples, i.e. is a white-box attack and needs full access to the model architecture and weights. It maximizes the model's loss $\mJ$ w.r.t. the input image via gradient ascent to create an adversarial image $\mX^{adv}$:
        \begin{equation*}
            \mX^{adv} = \mX +  \epsilon\cdot \text{sign}( \nabla_{\mX} \mJ(\mX_{N}^{adv},y_{t}))\;\text{,}
        \end{equation*}
     where $X$ is the benign input image, $y$ is the image label, and $\epsilon$ is a small scalar that ensures the perturbations are small.\\ 
     
     \noindent \textbf{\acf{bim}\,} \cite{bim} is an improved, iterative version of \ac{fgsm}. After each iteration the pixel values are clipped to the $\epsilon$ ball around the input image (i.e. $[x-\epsilon, x+\epsilon]$) as well as the input space (i.e. $[0, 255]$ for the pixel values):
        \begin{equation*}
            \begin{aligned}
                \mX_{0}^{adv}   &= \mX, \qquad  \\
                \mX_{N+1}^{adv} &= \text{CLIP}_{\mX,\epsilon} \{ \mX_{N}^{adv} + \alpha\cdot\text{sign}( \nabla_{\mX} \mJ(\mX,y_{t})) \},
            \end{aligned}
        \end{equation*}
        for iteration $N$ with step size $\alpha$.\\
     
     \noindent\textbf{\acf{pgd}\,} \label{sec:pgd}
         \cite{pgd}  is similar to \ac{bim} and one of the currently most popular 
        attacks. 
        PGD adds random initializations of the perturbations for each iteration. 
        Optimized perturbations are again projected onto the $\epsilon$ ball to ensure the similarity between the original and attacked image in terms of $L^2$ or $L^{\infty}$ norm.\\ 
        
    \noindent\textbf{\acf{autoattack}\,} 
    \cite{autoattack} is an ensemble of four parameter-free attacks: two parameter-free variants of \ac{pgd} \cite{pgd} using cross-entropy loss in \apgdce~ and difference of logits ratio loss (DLR) in \apgdt:
        \begin{equation}
            \text{DLR}(\vx,
            \vy) = \frac{z_y - \max_{\vx\neq \vy} z_i}{z_{\pi 1} - z_{\pi 3} }.
        \end{equation}
        where $\pi$ is the ordering of the components of $z$ in decreasing order. 
        Further \ac{autoattack} comprises a targeted version of the FAB attack \cite{fabtattack}, and the \squaredef~ attack \cite{squareattack} which is a black-box attack. 
        In \textit{RobustBench}, models are evaluated using \ac{autoattack}\,  in the standard mode where the four attacks are executed consecutively. If a sample's prediction can not be flipped by one attack, it is handed over to the next attack method, to maximize the overall attack success rate.\\
     
     \noindent \textbf{\acf{df}\,}  is a  non-targeted attack that finds the minimal amount of perturbation required to flip the network decision by an iterative linearization approach \cite{deepfool}. It thus estimates the distance from the input sample to the model decision boundary.\\
     
     \noindent\textbf{\ac{cw}\,} \cite{cw} uses a direct numerical optimization of inputs $X^{adv}$ such as to flip the network's prediction at the minimum required perturbation 
     and provides results optimized with respect to $L^2$, $L^0$ and $L^{\infty}$ distances. 
     In our evaluation, we use the $L^2$ distance for \ac{cw}.\\

\noindent\textbf{Adversarial Training\,}
 denotes the concept of using adversarial examples to augment the training data of a neural network. Ideally, this procedure should lead to better and denser coverage of the latent space and thus increased model robustness. \ac{fgsm} \cite{fgsm} adversarial training offers the advantage of rather fast adversarial training data generation. Yet, models tend to overfit to the specific attack such that additional tricks like early stopping \cite{rice2020overfitting,wong2020fast} have to be employed. Training on multi-step adversaries generalizes more easily, yet is hardly affordable for large-scale problems such as \imagenet \, due to its computation costs.

\subsection{Adversarial Detection}
Adversarial Detection aims to distinguish adversarial examples from benign examples and is thus a low computational replacement to an expensive adversarial training strategy. In test scenarios, adversarial attacks can be rejected and cause faulty classifications.

\noindent Given a trained DNN  on a clean dataset for the origin task, many existing methods \cite{ma2018characterizing, feinman,  mah, original, lorenz2021detecting} train a binary classifier on top of some hidden-layer embeddings of the given network as the adversarial detector. 
The strategy is motivated by the observation that adversarial examples have very different distributions from natural examples of intermediate-layer features. So a detector can be built upon some statistics of the distribution, i.e., \ac{kd} \cite{feinman}, \ac{mah} \cite{mah} distance, or \ac{lid} \cite{ma2018characterizing}.
SpectralDefense (SD) approaches (blackbox and whitebox) \cite{original, lorenz2021detecting, lorenz2022is} aim to detect adversarial images by their frequency spectra in the input or feature map representation.

\noindent Complementary, \cite{cdvae} propose to train a variational autoencoder following the principle of the class distanglement. They argue that the reconstructions of adversarial images are characteristically different and can more easily be detected using for example \ac{kd}, \ac{mah} and \ac{lid}).\\

\noindent\textbf{Local Intrinsic Dimensionality (LID)} is a measure that represents the average distance from a point to its neighbors in a learned representation space \cite{amsaleg,houle17a} and thereby approximates the intrinsic dimensionality of the representation space via maximum likelihood estimation. 

\noindent Let $\mathcal{B}$ be a mini-batch of $N$ clean examples and Let $r_i(\vx)=d(\vx,\vy)$  be the Euclidean distance between the sample $\vx$ and its i-th nearest neighbor in  $\mathcal{B}$. Then, the LID can be approximated by 
\begin{equation}
    \text{LID}(\vx) = - \left (\frac{1}{k} \sum^k_{i=1} \log \frac{d_i(\vx)}{d_k(\vx)}  \right )^{-1}, \label{eq:lid}
\end{equation}
where $k$ is a hyper-parameter that controls the number of nearest neighbors to consider, and $d$ is the employed distance metric.
Ma \etal \cite{ma2018characterizing} propose to use LID to characterize properties of adversarial examples, i.e.~they argue that the average distance of samples to their neighbors in the learned latent space of a classifier is characteristic of adversarial and benign samples. Specifically, they evaluate LID for the $j$-dimensional latent representations of a neural network $f(\vx)$ of a sample $\vx$ use the $L^2$ distance 
\begin{equation}
    d_{\ell}(\vx,\vy) = \| f_{\ell}^{1..j}(\vx) - f_{\ell}^{1..j}(\vy) \|_2
\end{equation} 
for all $\ell \in L$ feature maps. They compute a vector of LID values for each sample:
\begin{equation}
    \overrightarrow{\mathrm{LID}}(\vx) = \{ \mathrm{LID}_{d_\ell} (\vx) \}^n_\ell. \label{eq:featuremaps}
\end{equation}
Finally, they compute the $\overrightarrow{\mathrm{LID}}(\vx)$ over the training data and adversarial examples generated on the training data, and train a logistic regression classifier to detect adversarial\footnote{We are grateful to the authors for releasing their complete source code.  \url{https://github.com/xingjunm/lid_adversarial_subspace_detection}.} samples. 

\section{Revisiting Local Intrinsic Dimensinality}\label{sec:revisiting}
The LID method for adversarial example detection as proposed in \cite{ma2018characterizing} was motivated by the MLE estimate for the intrinsic dimension as proposed by \cite{amsaleg}. We refer to this original formulation to motivate our proposed multiLID. Let us denote $\mathbb{R}^m,d$ a continuous domain with non-negative distance function $d$. The continuous intrinsic dimensionality aims to measure the local intrinsic dimensionality of $\mathbb{R}^m$ in terms of the distribution of inter-point distances. Thus, we consider for a fixed point $\vx$ the distribution of distances as a random variable $\mathbf{D}$ on $[o,+\infty)$ with probability density function $f_D$ and cumulative density function $F_D$. 
For samples $\vx$ drawn from continuous probability distributions, the intrinsic dimensionality is then defined as in \cite{amsaleg}:\\

\begin{definition}{Instrinsic Dimensionality (ID).}
Given a sample $x\in \mathbb{R}^m$, let $D$ be a random variable denoting the distance from $x$ to other data samples. If the cumulative distribution $F(d)$ of $\mathbf{D}$ is positive and continuously differentiable at distance $d>0$, the ID of x at distance d is given by:
\begin{equation}
    \mathrm{ID}_{\mathbf{D}}(d) \overset{\Delta}{=} \text{lim}_{\epsilon\rightarrow 0}\frac{\mathrm{log} F_\mathbf{D}((1+\epsilon)d) - \mathrm{log} F_\mathbf{D}(d)}{\mathrm{log}(1+\epsilon)} 
    \label{eq:id_def}
\end{equation}
\end{definition}
\noindent In practice, we are given a fixed number $n$ of samples of $\vx$ such that we can compute their distances to $\vx$ in ascending order $d_1\leq d_2\leq \dots \leq d_{n-1}$ with maximum distance $w$ between any two samples. As shown in \cite{amsaleg}, the log-likelihood of $\text{ID}_{\mathbf{D}}(d)$ for $\vx$ is then given as
\begin{equation}
n\mathrm{log}\frac{F_{\mathbf{D},w}(w)}{w} + n\mathrm{log}\mathrm{ID}_\mathbf{D} + (\mathrm{ID}_\mathbf{D}-1)\sum_{i=1}^{n-1} \mathrm{log}\frac{d_i}{w}.
\end{equation}
The maximum likelihood estimate is then given as 
\begin{equation}
    \widehat{\mathrm{ID}}_{\mathbf{D}} = -\left(\frac{1}{n}\sum_{i=0}^{n-1}\mathrm{log}\frac{d_i}{w}\right)^{-1}\qquad \mathrm{with} \label{MLE}
\end{equation}

\begin{equation}
    \widehat{\mathrm{ID}}_\mathbf{D}\sim \mathcal{N}\left(\mathrm{ID}_\mathbf{D},\frac{ \mathrm{ID}_\mathbf{D}^2}{n}\right),
\end{equation}
i.e. the estimate is drawn from a normal distribution with mean $\mathrm{ID}_\mathbf{D}$ and its variance decreases linearly with an increasing number of samples while it increases quadratically with $\mathrm{ID}_\mathbf{D}$. The \emph{local} ID is then an estimate of the ID based on the local neighborhood of $\vx$, for example, based on its $k$ nearest neighbors. This corresponds to equation \eqref{eq:lid}. 
This local approximation has the advantage of allowing for an efficient computation even on a per batch basis as done in \cite{ma2018characterizing}. It has the disadvantage that it does not consider the strong variations in variances $\mathrm{ID}_\mathbf{D}^2/n$, i.e.~the estimates might become arbitrarily poor for large $\mathrm{ID}$ if the number of samples is limited. This becomes even more severe as \cite{amsaleg_21} showed that latent representations with large $\mathrm{ID}$ are particularly vulnerable to adversarial attacks.

\begin{figure*}[h]
    \centering
    \includegraphics[angle=0, width=1\textwidth]{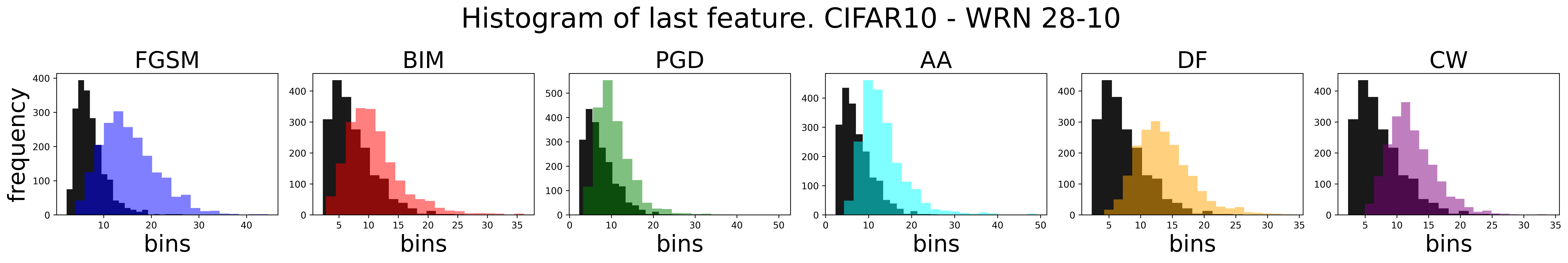}
    \caption{
    \corrected{
    Visualization of the LID features from the clean set of samples (black) and different adversarial attacks of 100 samples. 
    The network is \wideresnetcif~ trained on \cifar and LID is evaluated on the feature map after the last ReLU activation.  
    }
    }
    \label{fig:lastlinefeatures}
\end{figure*}

\noindent In \cref{fig:lastlinefeatures}, we evaluate the distribution of LID estimates computed for benign and adversarial examples of different attacks on the latent feature representation of a classifier network (see \cref{experiments}). We make the following two observations: (i) the distribution has a rather long tail and is not uni-modal, i.e.~we are likely to face rather strong variations in the ID for different latent sub-spaces, and (ii) the LID estimates for adversarial examples have the tendency to be higher than the ones for benign examples, (iii) the LID is more informative for some attacks and less informative on others.
As a first conclusion, we expect the discrimination between adversarial examples and benign ones to be particularly hard when the tail of the distribution is concerned, i.e. for those benign points with rather large LID that can only be measured at very low confidence according to equation \eqref{MLE}. Secondly, we expect linear separation methods based on LID such as suggested by \cite{ma2018characterizing} to be unnecessarily weak, and third, we expect the choice of the considered layers to have a rather strong influence on the expressiveness of LID for adversarial detection.

\noindent As a remedy, we propose several rather simple improvements:
\begin{itemize}
\item We propose to unfold the aggregated LID estimates in equation \eqref{eq:lid} and rather consider the normalized log distances between a sample and its neighbors separately in a feature vector, which we denote \emph{multiLID}.
\item We argue that the deep network layers considered to compute LID or multiLID have to be carefully chosen. An arbitrary choice might yield poor results. 
\item Instead of using a logistic regression classifier, highly non-linear classifiers such as a random forest should increase LID-based discrimination between adversarial and benign samples.

\end{itemize}

\noindent Let us analyze the implications of the LID unfolding in more detail. As argued for example in \cite{ma2018characterizing} before, the empirically computed LID can be interpreted as an estimate of the local growth rate similar to previous generalized expansion models \cite{ruhl,Houle12}. Thereby, the idea is to deduce the expansion dimension from the volume growth around a sample, and the growth rate is estimated by considering probability mass in increasing distances from the sample. Such expansion models, like the LID, are estimated within a local neighborhood around each sample and therefore provide a local view of the data dimensionality \cite{ma2018characterizing}. The local ID estimation in eq.~\eqref{eq:lid} can be seen as a statistical interpretation of a growth rate estimate. Please refer to \cite{houle17a,houle17}  for more details.

\begin{figure*}[h]
    \centering
    \includegraphics[angle=0, width=0.99\textwidth]{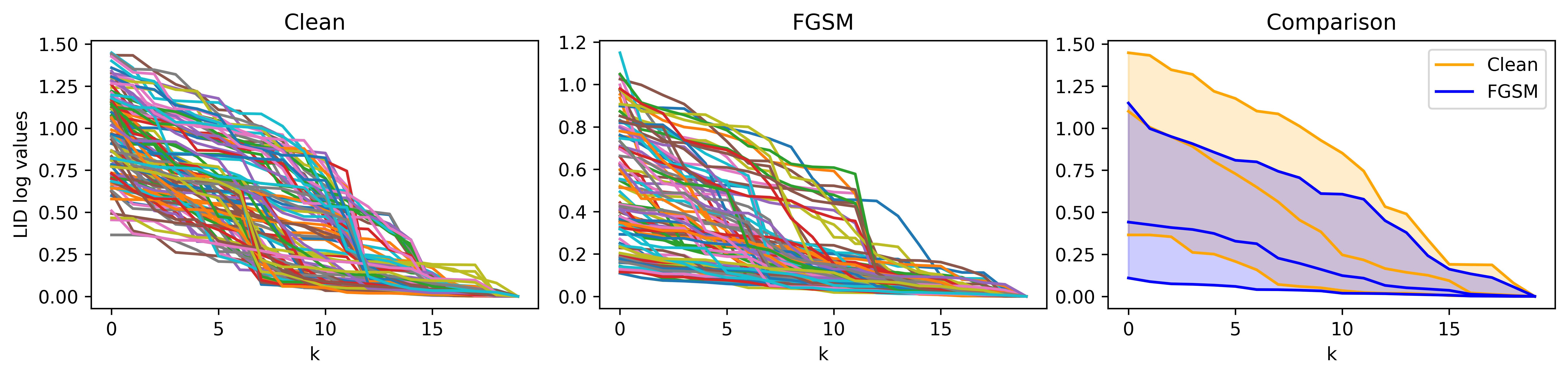}
    \caption{
    \corrected{
    Visualization of the LID features from the clean and \ac{fgsm} set of 100 samples over each $k$. 
    The network is \wideresnetcif~ trained on \cifar. 
    The feature values for the nearest neighbors (low values on the x-axis) are significantly higher for the clean dataset. 
    The LID log values are inversely proportional to the distance as shown in  \cref{eq:multilid}.
    The plot on the right illustrates the mean and standard deviation of the two sets of profiles.
    }
    }
    \label{fig:logvis_fgsm}
\end{figure*}

\noindent In practical settings, this statistical estimate not only depends on the considered neighborhood size. LID is usually evaluated on a mini-batch basis, i.e.~the $k$ nearest neighbors are determined within a random sample of points in the latent space. While this setting is necessarily relatively noisy, it offers a larger coverage of the space while considering only a few neighbors in every LID evaluation. Specifically, the relative growth rate is aggregated over potentially large distances within the latent space, when executing the summation in eq.~\eqref{eq:lid}. We argue that this summation step integrates potentially very discriminative information since it mixes local information about the growth rate in direct proximity with more distantly computed growth rates. Therefore, we propose to "unfold" this growth rate estimation. Instead of the aggregated (semi) local ID, we propose to compute for every sample $\vx$ a feature vector, denoted \emph{multiLID}, with length $k$ as
\begin{equation}
   \overrightarrow{\text{multiLID}_d(\vx)}[i] = - \left ( \log \frac{d_i(\vx)}{d_k(\vx)}  \right )^{-1}. \label{eq:multilid}
\end{equation}
where $d$ is measured using the Euclidean distance. Figure \ref{fig:logvis_fgsm} visualizes multiLID for 100 benign CIFAR10 samples and samples that have been perturbed using \ac{fgsm}. It can easily be seen that several characteristic profiles in the multiLID would be integrated into very similar LID estimates while being discriminative when all $k$ growth ratio samples are considered as a vector. MultiLID facilitates to leverage of the different characteristic growth rate profiles.


\section{Experiments}  \label{experiments}
    
     To validate our proposed multiLID, we conduct extensive experiments on \cifar, \cifarhun, and \imagenet.  We train two different models, a wide-resnet (WRN 28-10) \cite{wideresidual, wider} and a VGG-16 model \cite{vgg} on the different datasets. While we use test samples from the original datasets as clean samples, we generate adversarial samples using a variety of adversarial attacks. 
     From clean and adversarial data, we extract the feature maps for different layers, at the output of the ReLU activations. We use a random subset of 2000 samples of this data for each attack method and extract the multiLID features from the feature maps. From this random subset, we take a train-test split of 80:20, i.e. we have a training set of 3200 samples (1600 clean, 1600 attacked images) and a balanced test set of 400 images for each attack. This setting is common practice as used in \cite{lid, mah, lorenz2022is}. All experiments were conducted on  3 Nvidia A100 40GB GPUs for \imagenet~ and 3 Nvidia Titan with 12GB for \cifar~ and \cifarhun.\\
 
    \noindent\textbf{Datasets. } Many of the adversarial training methods ranked on Robustbench\footnote{\url{https://robustbench.github.io}} are based on the \wideresnetcif~ \cite{wideresidual, wider} architecture. Therefore, we also conduct our evaluation on a baseline \wideresnetcif~and train it with clean examples.  \\
    \textit{\cifar:} The \cifar~ \wideresnetcif~reaches a test accuracy of 96\% and the VGG-16 model reaches  72\% top-1 accuracy \cite{lorenz2022is} on the test set. We then apply the different attacks on the test set.  \\
    \textit{\cifarhun:} 
    The procedure is equal to \cifar~dataset. We report a test-accuracy for \wideresnetcif~ of 83\% (VGG-16 reaches 81\%) \cite{lorenz2022is} . 
    \\ \textit{\imagenet:}
      The PyTorch library provides a pre-trained \wideresnetim~\cite{wideresidual} for ImageNet. As a test set, we use the official validation set from \imagenet~and reach a validation accuracy of 80\%.\\  

    \noindent\textbf{Attack methods.} \label{sec:data_generation}
         We generate test data from the six most commonly used adversarial attacks: \ac{fgsm}, \ac{bim}, \ac{pgd}(-$L^{\infty}$), \ac{cw}(-$L^{2}$), \ac{df}(-$L^{2}$) and \ac{autoattack}, as explained in  \cref{sec:adversarial_attacks}. For  \ac{fgsm}, \ac{bim}, \ac{pgd}(-$L^{\infty}$), and \ac{autoattack}, we use the commonly employed perturbation size of $\epsilon=8/255$, \ac{df} is limited to 20 iterations and \ac{cw} to 1000 iterations. \\

    \noindent\textbf{Layer feature selection per architecture.}  \label{sec:featureselection}
    Following \cref{eq:featuremaps}, for the \textit{\wideresnetcif} and \textit{\wideresnetim}, we focus on the ReLU activation layers, whereas in each residual block, we take the last one. This results in 13 activations layers for \textit{\wideresnetcif} and 17 for \textit{\wideresnetim} to compute multiLID representations. This is different from the setting proposed in \cite{cdvae}, which propose to use the outputs of the three convolutional blocks. In \cite{ma2018characterizing} only simpler network architectures have been considered and the feature maps at the output of every layer are considered to compute LID.
    For the VGG-16 architecture, according to \cite{original}, we take the features of all activation layers, which are again 13 layers in total.\\
    
    \noindent\textbf{Minibatch size in LID estimation.} \label{sec:minibacht}
    As motivated in \cite{ma2018characterizing}, we estimate the multiLID values using a default minibatch size $|\mathcal{B}|$ of 100 with $k$ selected as of 20\%  of mini-batch size \cite{ma2018characterizing}.  As discussed above and theoretically argued before in \cite{amsaleg} the MLE estimator of LID suffers on such small samples, yet, already provides reasonable results when used for adversarial detection \cite{ma2018characterizing}. Our proposed multiLID can perform very well in this computationally affordable setting across all datasets.

\subsection{Results}

In this section, we report the final results of our \textit{multi LID} method and compare it to competing mapproaches  In \cref{tab:reslid}, we compare the results of the \textit{original LID} \cite{ma2018characterizing} to the results of our proposed \textit{multiLID} method for both model types, the wide-resnets and VGG-16 models on the three datasets \cifar, \cifarhun, and \imagenet. For LID and the proposed multiLID, we extract features from the same layers in the network to facilitate direct comparison. While LID already achieves overall good results the proposed multiLID can even perfectly discriminate between benign and adversarial images on these data in terms of AUC as well as F1 score.

\begin{table*}[th]
\centering
\caption{
Results. Comparison of the original LID method with our proposed multiLID on different datasets. 
We report the AUC and F1 score as 
mean and variance over three evaluations with randomly drawn test samples. \label{tab:reslid} 
} 
\corrected{
\scalebox{0.67}{
    \begin{tabular}{lrrrrrrrrrr}
    \toprule[1pt]
    \midrule
    \multirow{3}{*}{\textbf{Attacks}} & \multicolumn{4}{c}{\textbf{CIFAR10}} & \multicolumn{4}{c}{\textbf{CIFAR100}} & \multicolumn{2}{c}{\textbf{ImageNet}} \\  \cmidrule{2-11}  
     & \multicolumn{2}{c}{\textbf{WRN 28-10}} & \multicolumn{2}{c}{\textbf{VGG16}} & \multicolumn{2}{c}{\textbf{WRN 28-10}} & \multicolumn{2}{c}{\textbf{VGG16}} & \multicolumn{2}{c}{\textbf{WRN 50-2}} \\ \cmidrule{2-11} 
     & \multicolumn{1}{c}{\textbf{AUC}} & \multicolumn{1}{c|}{\textbf{F1}} & \multicolumn{1}{c}{\textbf{AUC}} & \multicolumn{1}{c|}{\textbf{F1}} & \multicolumn{1}{c}{\textbf{AUC}} & \multicolumn{1}{c|}{\textbf{F1}} & \multicolumn{1}{c}{\textbf{AUC}} & \multicolumn{1}{c|}{\textbf{F1}} & \multicolumn{1}{c}{\textbf{AUC}} & \multicolumn{1}{c}{\textbf{F1}} \\ \midrule
    \multicolumn{11}{c}{\textbf{ original LID \cite{ma2018characterizing}}} \\ \midrule
    \textbf{FGSM} & $95.89 \pm 0.07$ & $89.46 \pm 0.01$ & $87.76 \pm 0.21$ & $78.58 \pm 0.15$ & $97.71 \pm 0.82$ & $92.71 \pm 2.17$ & $77.20 \pm 1.48$ & $70.08 \pm 0.31$ & $71.40 \pm 9.07$ & $65.96 \pm 4.34$ \\
    \textbf{BIM} & $86.68 \pm 0.50$ & $78.50 \pm 0.46$ & $87.73 \pm 0.68$ & $78.53 \pm 0.52$ & $95.53 \pm 1.02$ & $87.75 \pm 0.93$ & $81.31 \pm 3.62$ & $73.73 \pm 6.66$ & $94.02 \pm 0.27$ & $86.58 \pm 0.50$ \\
    \textbf{PGD} & $88.92 \pm 0.86$ & $80.11 \pm 1.77$ & $84.78 \pm 0.68$ & $74.93 \pm 1.86$ & $97.76 \pm 0.06$ & $91.40 \pm 0.09$ & $84.75 \pm 1.59$ & $78.04 \pm 2.47$ & $95.81 \pm 1.00$ & $88.54 \pm 3.08$ \\
    \textbf{AA} & $96.49 \pm 0.32$ & $90.78 \pm 0.22$ & $95.25 \pm 0.49$ & $87.35 \pm 2.30$ & $99.18 \pm 0.03$ & $94.74 \pm 0.85$ & $87.02 \pm 0.83$ & $78.81 \pm 0.23$ & $99.87 \pm 0.00$ & $98.01 \pm 0.06$ \\
    \textbf{DF} & $94.40 \pm 0.07$ & $86.08 \pm 3.15$ & $85.93 \pm 0.26$ & $75.88 \pm 0.74$ & $57.02 \pm 0.39$ & $52.59 \pm 1.44$ & $54.39 \pm 0.03$ & $52.93 \pm 0.32$ & $54.62 \pm 0.07$ & $49.04 \pm 2.06$ \\
    \textbf{CW} & $92.83 \pm 0.21$ & $84.33 \pm 1.50$ & $83.34 \pm 0.47$ & $74.51 \pm 0.53$ & $55.07 \pm 0.31$ & $53.74 \pm 4.31$ & $61.47 \pm 0.86$ & $61.99 \pm 2.12$ & $54.45 \pm 0.05$ & $50.36 \pm 3.86$ \\
    \midrule
    \multicolumn{11}{c}{multiLID + improved layer setting + RF or short: \textbf{ multiLID   (ours) }} \\ \midrule
    \textbf{FGSM} & $96.98 \pm 0.20$ & $91.19 \pm 0.95$ & $90.78 \pm 0.38$ & $82.34 \pm 1.37$ & $98.55 \pm 0.28$ & $94.80 \pm 1.32$ & $83.00 \pm 0.84$ & $76.89 \pm 0.02$ & $79.25 \pm 2.75$ & $72.98 \pm 0.67$ \\
    \textbf{BIM} & $96.10 \pm 0.39$ & $89.93 \pm 1.24$ & $94.50 \pm 0.26$ & $88.14 \pm 0.43$ & $97.88 \pm 0.07$ & $91.72 \pm 0.09$ & $82.96 \pm 0.94$ & $75.42 \pm 1.39$ & $94.48 \pm 0.18$ & $86.92 \pm 0.93$ \\
    \textbf{PGD} & $97.69 \pm 0.12$ & $92.81 \pm 0.37$ & $92.35 \pm 1.46$ & $83.52 \pm 3.46$ & $98.76 \pm 0.05$ & $94.74 \pm 1.14$ & $88.39 \pm 0.06$ & $81.42 \pm 0.44$ & $96.39 \pm 0.11$ & $90.21 \pm 0.49$ \\
    \textbf{AA} & $99.45 \pm 0.03$ & $96.88 \pm 0.00$ & $98.77 \pm 0.08$ & $94.85 \pm 1.42$ & $99.85 \pm 0.00$ & $98.33 \pm 0.01$ & $91.25 \pm 0.33$ & $83.48 \pm 0.17$ & $99.90 \pm 0.00$ & $98.83 \pm 0.04$ \\
    \textbf{DF} & $97.51 \pm 0.12$ & $94.04 \pm 0.26$ & $89.37 \pm 2.42$ & $84.32 \pm 0.83$ & $74.75 \pm 0.16$ & $70.18 \pm 0.57$ & $73.78 \pm 1.12$ & $70.04 \pm 0.09$ & $52.93 \pm 1.01$ & $52.77 \pm 1.89$ \\
    \textbf{CW} & $97.92 \pm 0.01$ & $96.00 \pm 0.11$ & $89.75 \pm 0.21$ & $85.27 \pm 0.85$ & $70.10 \pm 1.33$ & $67.77 \pm 0.85$ & $76.15 \pm 0.20$ & $71.59 \pm 0.20$ & $53.37 \pm 0.01$ & $52.04 \pm 0.17$ \\
	\midrule
	\bottomrule[1pt]
    \end{tabular}
    }
}
\end{table*}

\begin{table*}[h]
\centering
\caption{Comparison of multiLID with the state-of-the-art on \cifar.} \label{tab:comparativecif10}
\scalebox{0.7}{
    \begin{tabular}{lrrrrrrrr}
    \toprule[1pt] \midrule
    \multicolumn{9}{c}{\textbf{\cifar \, on \wideresnetcif}} \\ \midrule
    \multirow{2}{*}{\textbf{Defenses}} & \multicolumn{2}{c}{\textbf{FGSM}} & \multicolumn{2}{c}{\textbf{BIM}} & \multicolumn{2}{c}{\textbf{PGD}} & \multicolumn{2}{c}{\textbf{CW}} \\ \cmidrule{2-9} 
     & \multicolumn{1}{c}{\textbf{TNR}} & \multicolumn{1}{c}{\textbf{AUC}} & \multicolumn{1}{c}{\textbf{TNR}} & \multicolumn{1}{c}{\textbf{AUC}} & \multicolumn{1}{c}{\textbf{TNR}} & \multicolumn{1}{c}{\textbf{AUC}} & \multicolumn{1}{c}{\textbf{TNR}} & \multicolumn{1}{c}{\textbf{AUC}} \\ \midrule   \textbf{Results reported by} \cite{cdvae} \\ \midrule
        KD & 42.38 & 85.74 & 74.54 & 94.82 & 73.12 & 94.59 & 73.33 & 94.75 \\
        KD (R($\vx$)) & 57.10 & 89.69 & 96.79 & 99.27 & 96.56 & 99.30 & 94.67 & 98.73 \\ \hline
        \acs{lid} & 69.05 & 93.60 & 77.73 & 95.20 & 71.52 & 93.19 & 74.98 & 94.32 \\
        \acs{lid} (R($\vx$)) & 92.60 & 98.59 & 86.42 & 97.29 & 87.54 & 97.57 & 76.42 & 95.10 \\ \hline
        \acs{mah} & 94.91 & 98.69 & 88.33 & 97.66 & 77.23 & 95.38 & 86.30 & 97.36 \\
        \acs{mah} (R($\vx$)) & 99.68 & 99.36 & 98.92 & 99.74 & 99.13 & 99.79 & 98.94 & 99.68 \\ 
        \midrule
        \textbf{Competing Methods} \\ \midrule
        \acs{mah} \cite{mah} & 97.37& 99.34&98.16&99.61&97.37&99.66&91.58&96.54 \\ 
        SpectralDefense$_\text{BlackBox}$ \cite{original} & 95.79 & 99.87 & 92.63 & 99.83 & 92.11 & 99.29 & 53.68 & 63.23 \\
        SpectralDefense$_\text{WhiteBox}$ \cite{original} & 99.47 & 100.00 & 96.32 & 99.99 & 95.79 & 99.97 & 84.47 & 96.89\\ 
         \textcolor{red}{LID,  settings from \cite{cdvae}} &  \textcolor{red}{87.25} &  \textcolor{red}{84.82} &  \textcolor{red}{85.02} &  \textcolor{red}{81.07} &  \textcolor{red}{81.61} &  \textcolor{red}{89.00} &  \textcolor{red}{85.89} &  \textcolor{red}{90.48}   \\
        \midrule
        \textbf{Ours} \\ \midrule
        
         \textcolor{red}{multiLID, settings from \cite{cdvae} + LR  }  &  \textcolor{red}{85.89} & \textcolor{red}{95.02} & \textcolor{red}{83.21} &  \textcolor{red}{95.56} &  \textcolor{red}{93.93} &  \textcolor{red}{98.00} &  \textcolor{red}{91.07} &  \textcolor{red}{97.05} \\
         
         \textcolor{red}{multiLID, settings from \cite{cdvae} + RF }   &  \textcolor{red}{87.50} &  \textcolor{red}{94.01} &  \textcolor{red}{85.89} &  \textcolor{red}{96.74} &  \textcolor{red}{94.64} &  \textcolor{red}{98.96} &  \textcolor{red}{92.14} &  \textcolor{red}{97.15} \\
         
         \textcolor{red}{LID, improved layer setting}                  &  \textcolor{red}{90.18} &  \textcolor{red}{96.62}  &  \textcolor{red}{ 93.21}   &  \textcolor{red}{98.18}  &  \textcolor{red}{85.89} &  \textcolor{red}{90.48} &  \textcolor{red}{87.50} &  \textcolor{red}{93.36} \\
         
         \rowcolor[rgb]{0.871,0.871,0.871}   \textcolor{red}{multiLID + improved layer setting + LR}  &  \textcolor{red}{90.18} & \textcolor{red}{ 96.62}   & \textcolor{red}{81.43} & \textcolor{red}{93.83} & \textcolor{red}{86.61} & \textcolor{red}{96.44} & \textcolor{red}{93.21} & \textcolor{red}{98.18} \\
         
        \rowcolor[rgb]{0.871,0.871,0.871} 
        \textcolor{red}{multiLID + improved layer setting + RF}  & \textcolor{red}{90.54} & \textcolor{red}{96.33}   & \textcolor{red}{85.71}  & \textcolor{red}{94.75}  & \textcolor{red}{94.64} & \textcolor{red}{98.96} & \textcolor{red}{92.14} & \textcolor{red}{97.15} \\
		\midrule
		\bottomrule[1pt]
    \end{tabular}
}
\end{table*}

\begin{table*}[h!]
\centering
\caption{Results of using multiLID. Comparison of \ac{lr} and \ac{rf} classifier on different datasets. Comparison to \cref{tab:reslid} which uses \acs{lr}. The minibatch size is $|\mathcal{B}|= 100$ and the number of neighbors $k = 20$ according to \cref{sec:minibacht}.  } \label{tab:reslidfeatures}
\corrected{
\scalebox{0.67}{
\begin{tabular}{lrrrrrrrrrr}
\toprule[1pt]
\midrule
\multirow{3}{*}{\textbf{Attacks}} & \multicolumn{4}{c}{\textbf{CIFAR10}} & \multicolumn{4}{c}{\textbf{CIFAR100}} & \multicolumn{2}{c}{\textbf{ImageNet}} \\  \cline{2-11} 
 & \multicolumn{2}{c}{\textbf{WRN 28-10}} & \multicolumn{2}{c}{\textbf{VGG16}} & \multicolumn{2}{c}{\textbf{WRN 28-10}} & \multicolumn{2}{c}{\textbf{VGG16}} & \multicolumn{2}{c}{\textbf{WRN 50-2}} \\ \cmidrule{2-11} 
 & \multicolumn{1}{c}{\textbf{AUC}} & \multicolumn{1}{c|}{\textbf{F1}} & \multicolumn{1}{c}{\textbf{AUC}} & \multicolumn{1}{c|}{\textbf{F1}} & \multicolumn{1}{c}{\textbf{AUC}} & \multicolumn{1}{c|}{\textbf{F1}} & \multicolumn{1}{c}{\textbf{AUC}} & \multicolumn{1}{c|}{\textbf{F1}} & \multicolumn{1}{c}{\textbf{AUC}} & \multicolumn{1}{c}{\textbf{F1}} \\ \midrule
\multicolumn{11}{c}{\textbf{   multiLID + LR (ours)}} \\ \midrule
    \textbf{FGSM} & $97.64 \pm 0.16$ & $91.58 \pm 1.40$ & $93.24 \pm 0.05$ & $86.39 \pm 1.23$ & $99.10 \pm 0.05$ & $95.06 \pm 0.33$ & $85.32 \pm 0.02$ & $78.41 \pm 0.84$ & $80.65 \pm 3.30$ & $73.15 \pm 3.08$ \\ 
    \textbf{BIM} & $95.44 \pm 0.36$ & $90.40 \pm 0.28$ & $93.59 \pm 0.88$ & $87.02 \pm 1.27$ & $98.29 \pm 0.04$ & $92.84 \pm 0.14$ & $84.76 \pm 0.72$ & $77.76 \pm 3.38$ & $96.78 \pm 0.02$ & $90.54 \pm 0.32$ \\ 
    \textbf{PGD} & $96.60 \pm 0.36$ & $91.87 \pm 1.34$ & $90.91 \pm 0.95$ & $82.81 \pm 0.56$ & $98.89 \pm 0.00$ & $95.26 \pm 0.35$ & $89.10 \pm 0.18$ & $82.24 \pm 0.09$ & $97.44 \pm 0.06$ & $92.28 \pm 0.01$ \\ 
    \textbf{AA} & $99.00 \pm 0.06$ & $95.74 \pm 0.09$ & $98.25 \pm 0.24$ & $93.70 \pm 0.87$ & $99.89 \pm 0.00$ & $98.27 \pm 0.04$ & $91.27 \pm 0.71$ & $83.66 \pm 0.74$ & $99.95 \pm 0.00$ & $98.76 \pm 0.04$ \\ 
    \textbf{DF} & $97.93 \pm 0.03$ & $94.06 \pm 0.04$ & $89.54 \pm 1.02$ & $84.34 \pm 1.02$ & $75.99 \pm 0.73$ & $70.77 \pm 1.79$ & $70.76 \pm 4.48$ & $67.15 \pm 2.55$ & $54.49 \pm 0.18$ & $53.06 \pm 0.76$ \\ 
    \textbf{CW} & $97.86 \pm 0.03$ & $94.88 \pm 0.23$ & $89.79 \pm 0.79$ & $84.34 \pm 1.21$ & $71.01 \pm 1.23$ & $66.03 \pm 0.09$ & $70.79 \pm 0.90$ & $67.75 \pm 2.29$ & $54.51 \pm 0.97$ & $53.87 \pm 0.66$ \\     
\midrule
\multicolumn{11}{c}{\textbf{multiLID  + RF (ours)}} \\ \midrule
    \textbf{FGSM} & $96.84 \pm 0.14$ & $90.90 \pm 0.70$ & $91.04 \pm 0.24$ & $82.97 \pm 1.32$ & $98.47 \pm 0.31$ & $94.31 \pm 2.28$ & $82.56 \pm 0.77$ & $76.61 \pm 1.21$ & $79.52 \pm 2.09$ & $72.85 \pm 0.78$ \\
    \textbf{BIM} & $96.11 \pm 0.28$ & $89.61 \pm 1.77$ & $94.55 \pm 0.33$ & $88.37 \pm 0.99$ & $97.74 \pm 0.12$ & $91.48 \pm 0.14$ & $83.14 \pm 0.29$ & $74.94 \pm 0.13$ & $94.33 \pm 0.26$ & $87.43 \pm 2.04$ \\
    \textbf{PGD} & $97.50 \pm 0.11$ & $92.39 \pm 0.13$ & $92.18 \pm 1.07$ & $84.00 \pm 2.62$ & $98.77 \pm 0.06$ & $94.82 \pm 1.08$ & $88.36 \pm 0.09$ & $81.40 \pm 0.50$ & $96.55 \pm 0.19$ & $90.79 \pm 0.21$ \\
    \textbf{AA} & $99.54 \pm 0.01$ & $96.96 \pm 0.05$ & $98.61 \pm 0.02$ & $94.74 \pm 1.41$ & $99.87 \pm 0.00$ & $98.20 \pm 0.02$ & $91.25 \pm 0.36$ & $83.76 \pm 0.87$ & $99.88 \pm 0.01$ & $99.08 \pm 0.01$ \\
    \textbf{DF} & $97.47 \pm 0.21$ & $94.20 \pm 0.12$ & $89.04 \pm 1.61$ & $84.43 \pm 1.11$ & $74.85 \pm 0.58$ & $70.11 \pm 1.71$ & $73.44 \pm 0.66$ & $69.09 \pm 0.96$ & $53.41 \pm 2.63$ & $52.41 \pm 1.36$ \\
    \textbf{CW} & $97.91 \pm 0.04$ & $95.76 \pm 0.20$ & $89.67 \pm 0.52$ & $85.32 \pm 0.49$ & $69.80 \pm 0.65$ & $66.61 \pm 0.03$ & $76.12 \pm 0.07$ & $72.66 \pm 0.38$ & $53.80 \pm 2.39$ & $52.26 \pm 3.19$ \\
\midrule
\bottomrule[1pt]
\end{tabular}
}
}
\end{table*}

\noindent In \cref{tab:comparativecif10}, we further compare the AUC and F1 score, for \cifar~ trained on \wideresnetcif~ to a set of most widely used adversarial defense methods. 
First, we list the results from \cite{cdvae} for the defenses \emph{kernel density} (KD), \acs{lid}, and \ac{mah} as baselines. According to \cite{cdvae}, KD does not show strong results across the attacks, \ac{lid} and \ac{mah} yield a better average performance in their setting. For completeness, we also report the results CD-VAE  \cite{cdvae} by showing  $R(\vx)$ (which is the reconstruction of a sample $\vx$ through a $\beta$ variational autoencoder ($\beta$-VAE)). Encoding in such a well-conditioned latent space can help adversarial detection, yet is also time-consuming and requires task-specific training of the $\beta$-VAE. 

\noindent Our results, when reproducing LID on the same network layers as \cite{cdvae}, are reported in the second block of \cref{tab:comparativecif10}. While we can not exactly reproduce the numbers from \cite{cdvae}, the resulting AUC and F1 scores are in the same order of magnitude and slightly better in some cases. In this setting, LID performs slightly worse than the competing methods \ac{mah} and SpectralDefense$_\text{BlackBox}$ and SpectralDefense$_\text{WhiteBox}$ \cite{original}.  

\noindent We ablate on our different changes towards the full multiLID in the third block. When replacing LID by the unfolded features as in eq.~\eqref{eq:multilid} we already achieve results above 98\% F1 score in all settings. Defending against BIM is the hardest. The next line ablates on the employed feature maps. When replacing the convolutional features used in \cite{cdvae}\footnote{Assumption of CD-VAE LID layers taken from \url{https://github.com/kai-wen-yang/CD-VAE/blob/a33b5070d5d936396d51c8c2e7dedd62351ee5b2/detection/models/wide_resnet.py\#L86}.} by the last ReLU outputs in every block, we observe a boost in performance even on the plain LID features. Combining these two leads to almost perfect results. Results for other datasets are in \cref{tab:reslidfeatures}. F1-scores and AUC scores of consistently 100\% can be reached when classifying, on this feature basis, using a random forest classifier instead of the logistic regression. We refer to this setting as {\it multiLID} in all other tables including \cref{tab:reslid}.
%
%


\section{Ablation Study}
  
    In this section, we give insights on the different factors affecting our approach. We investigate the importance of the activation maps the features are extracted from as well as the number of multiLID features that are needed to reach good classification performance. Ablation on the number of considered neighbors as well as on the attack strength in terms of $\epsilon$ is provided in the Appendix. 



\subsection{Impact of non-linear Classification} \label{sec:nonlinear}
    In this section, we compare the methods from the last two lines of \cref{tab:comparativecif10} in more detail and for all three datasets. The results are reported in \cref{tab:reslid}. While the simple \ac{lr} classifier already achieves very high AUC and F1 scores on multiple for all attacks and datasets, \ac{rf} can further push the performance to even 100\%.


\subsection{Feature Importance}
\begin{figure}[H]
        \centering
        \includegraphics[width=\columnwidth]{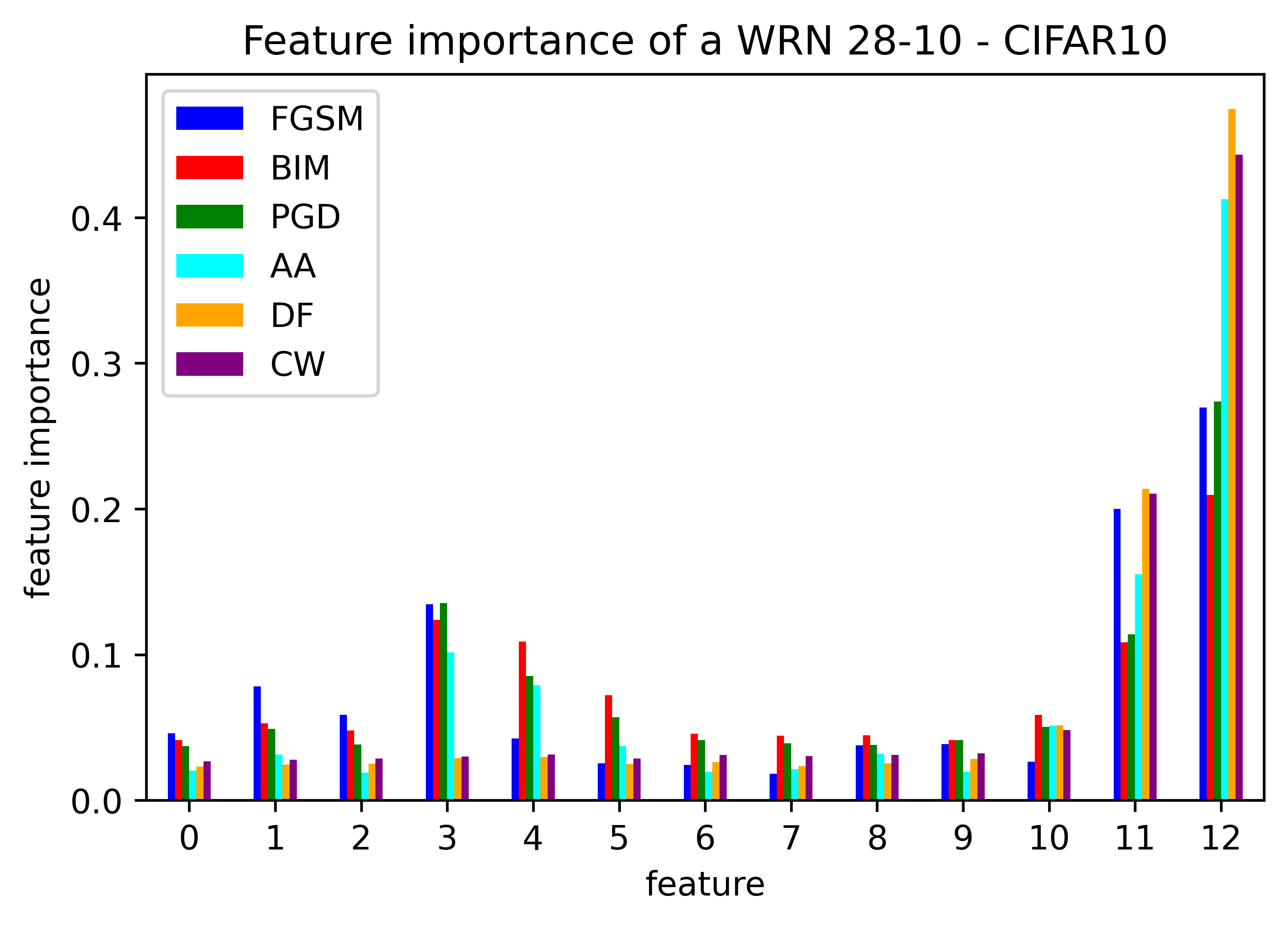}
        \caption{
        \corrected{
        Feature importance. Increasing order according to the activation function layers (feature) from \wideresnetcif~ trained on \cifar. The most relevant features are in the last ReLU layers. 
        }
        }
        \label{fig:featureimportance}
    \end{figure}

    The feature importance (variable importance) of the random forest describes the relevant features for the detection. In \cref{fig:featureimportance}, we plot the feature importance for the aggregated LID features of \wideresnetcif~ trained on a \cifar~dataset. The feature importance represents the importance of the selected ReLU layers (see \cite{mah}) in increasing order. The last features/layers show higher importance. For the attack \ac{fgsm}  the 3rd and last feature can be very relevant.
      \begin{figure}[tb]
        \centering
        \begin{subfigure}{.49\textwidth}
          \raggedright
          \includegraphics[scale=0.55]{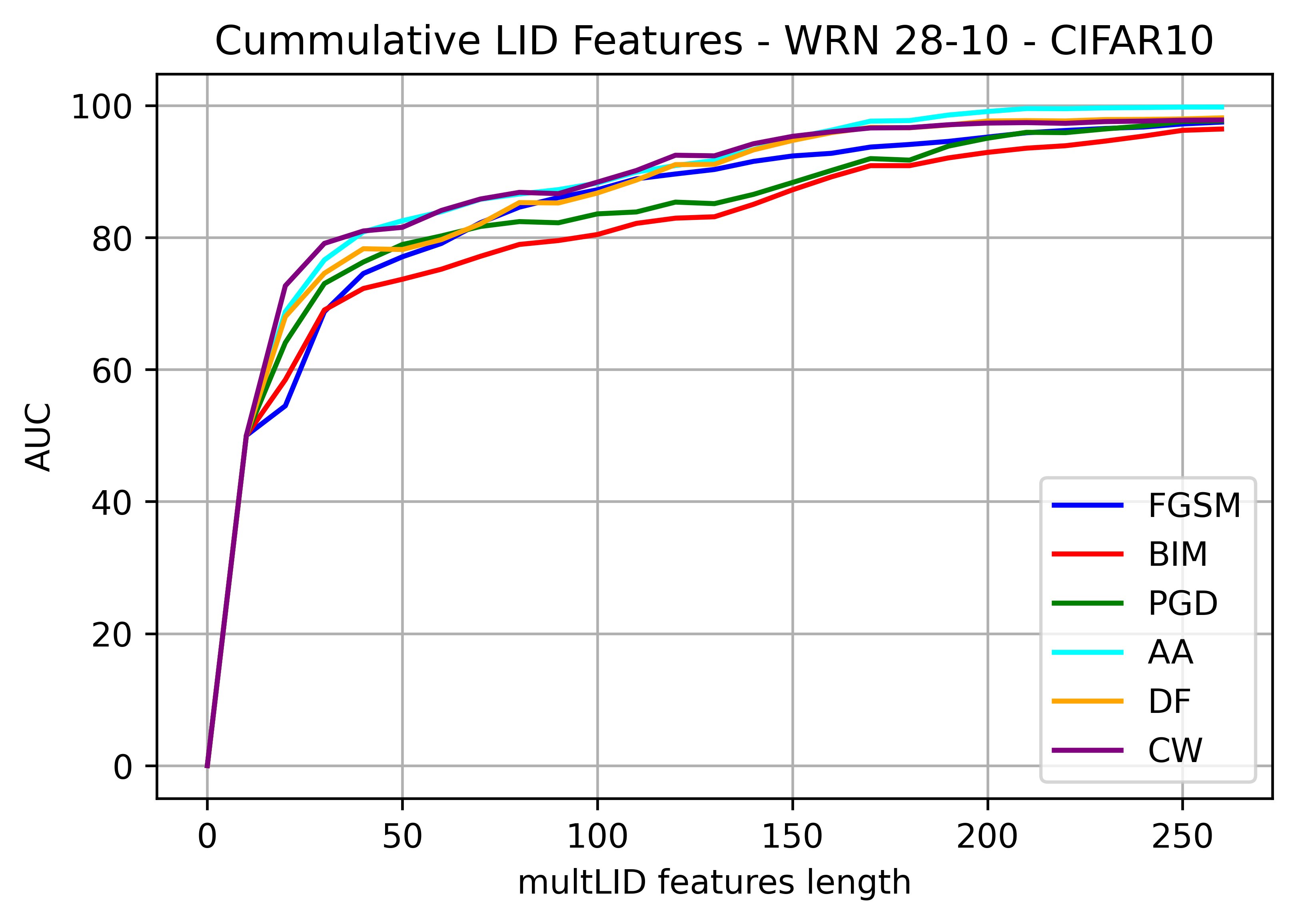}
          \caption{\corrected{Cumulative of all attacks on \cifar.}}
          \label{fig:sub1}
        \end{subfigure} \par\bigskip
        \begin{subfigure}{.49\textwidth}
          \raggedright
          \includegraphics[scale=0.55]{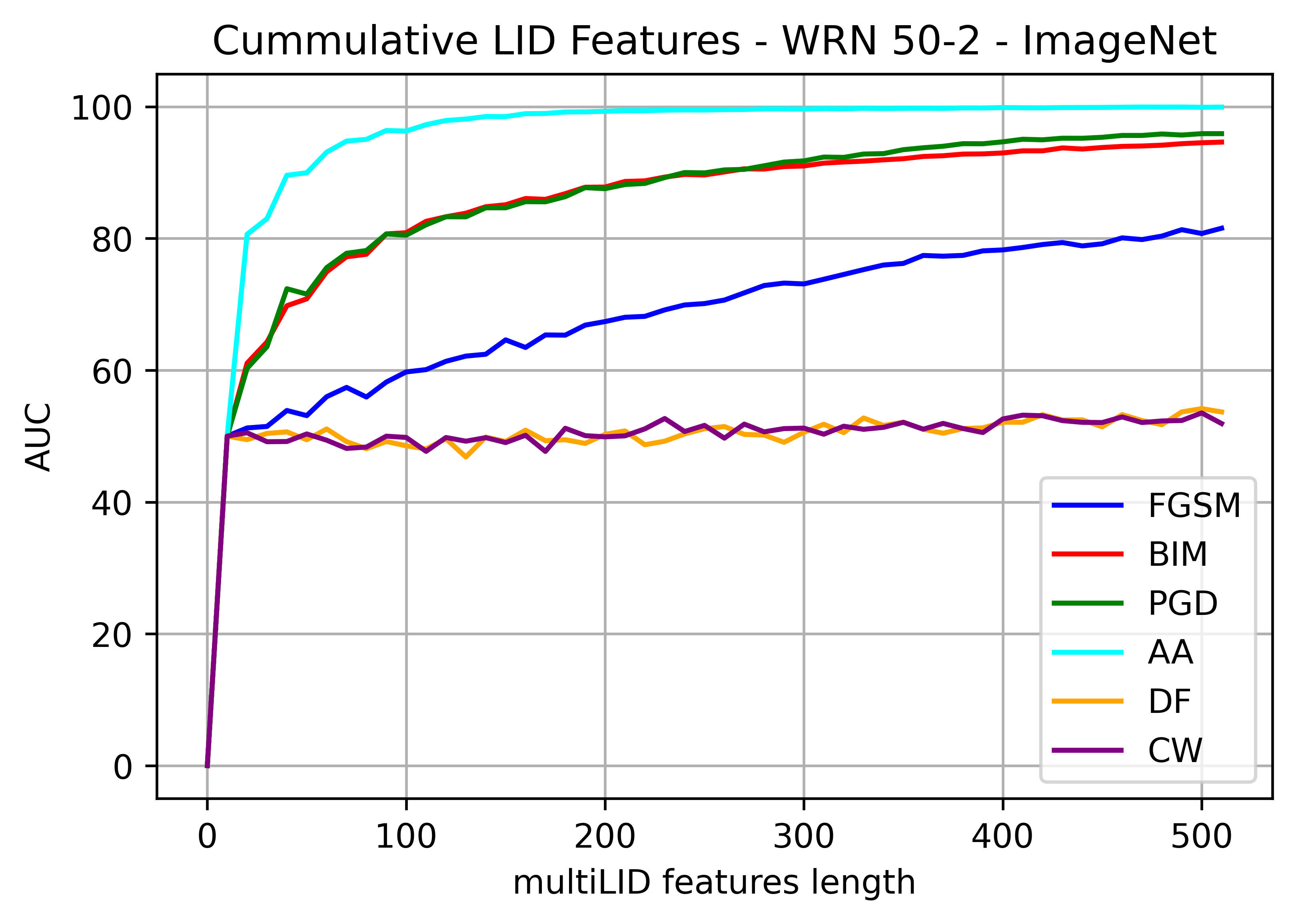}
          \caption{\corrected{Cumulative of all attacks on \imagenet.}}
          \label{fig:sub2}
        \end{subfigure}
        \caption{
        \corrected{
        Cumulative features used for the \ac{lr} classifier. The x-axis describes the length of the used feature vectors. The y-axis reports the AUC reached by using the most important features out of the full vector, sorted by \ac{rf} feature importance. 
        }
        }
        \label{fig:test}
    \end{figure}

\subsection{Investigation of the multiLID Features} \label{sec:lidfeatureslength}
        Following the \cref{eq:lid}, all neighbors $k$ are used for the classification. This time, we investigate the performance of the binary classifier logistic regression over the full multiLID features. For example, in \cref{fig:featureimportance} we consider 13 layers and the aggregated ID features for each. 
        Thus, the number of multiLID features per sample can be calculated as $\# \mathrm{layers} \times k$ which yields 260 features for $k=20$.
        In \cref{fig:test}, we visualize the AUC according to the length of the LID feature vectors, when successively more features are used according to their random forest feature importance. On \imagenet, it can be seen that \ac{df} and \ac{cw} need the full  length of these LID feature vectors to achieve the highest AUC scores. The observation, that the attacks \ac{df} and \ac{cw} are more effective are also reported in \cite{lorenz2022is}. Using a non-linear classifier on these very discriminant features, we can even achieve perfect F1 scores (see \cref{sec:nonlinear}).
   
   
   
   \subsection{Impact of the Number of Neighbors}
\begin{figure}[H]
    \begin{subfigure}[t]{.5\textwidth} \centering
      \includegraphics[scale=0.46]{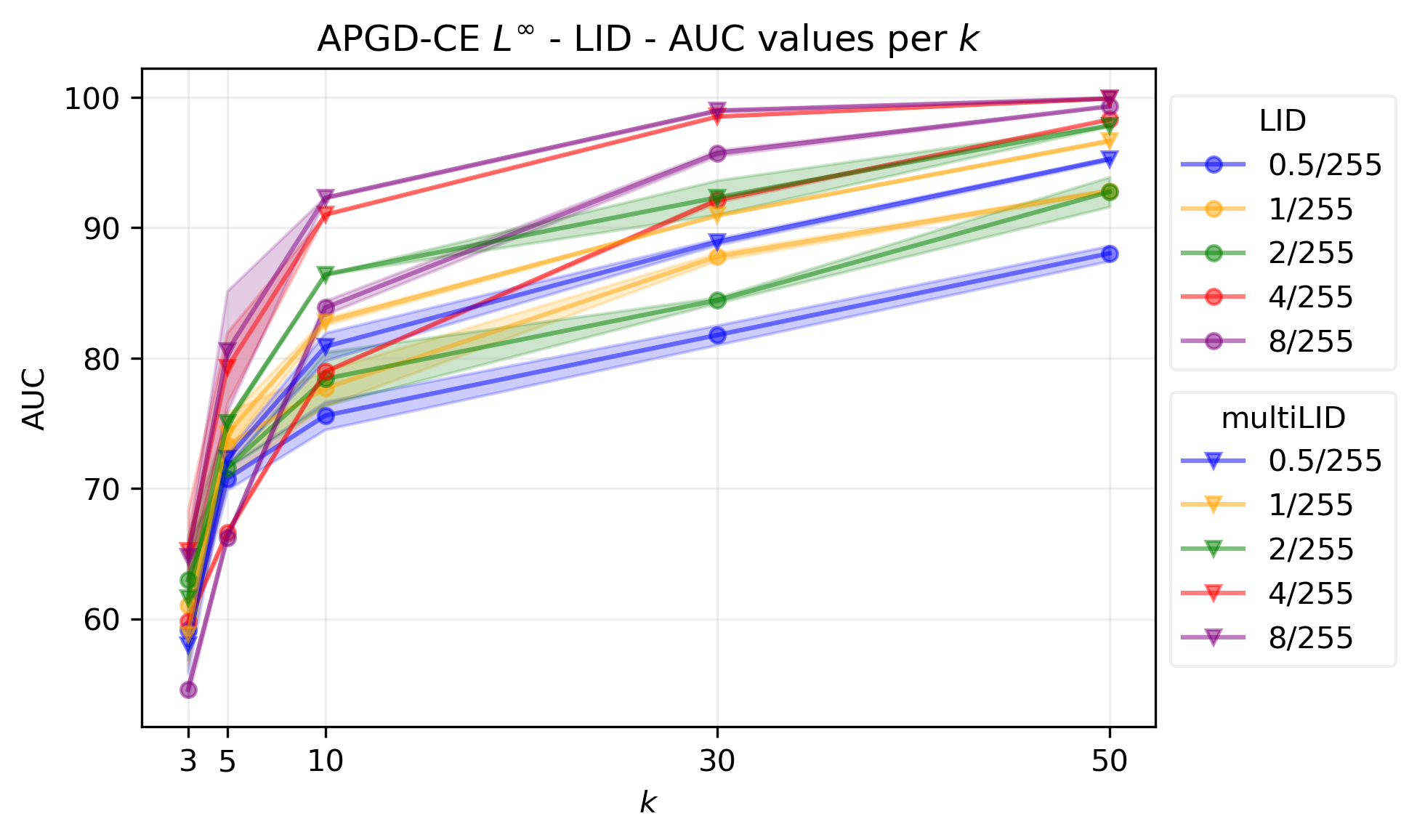}
      \caption{
      \corrected{
      The attack APGD-CE $L^{\infty}$ evaluated on different \\ epsilons and neighbors.
      }
      } \label{fig:multiinfeps}
    \end{subfigure} \par\bigskip
    \begin{subfigure}[t]{.5\textwidth} \centering
      \includegraphics[scale=0.46]{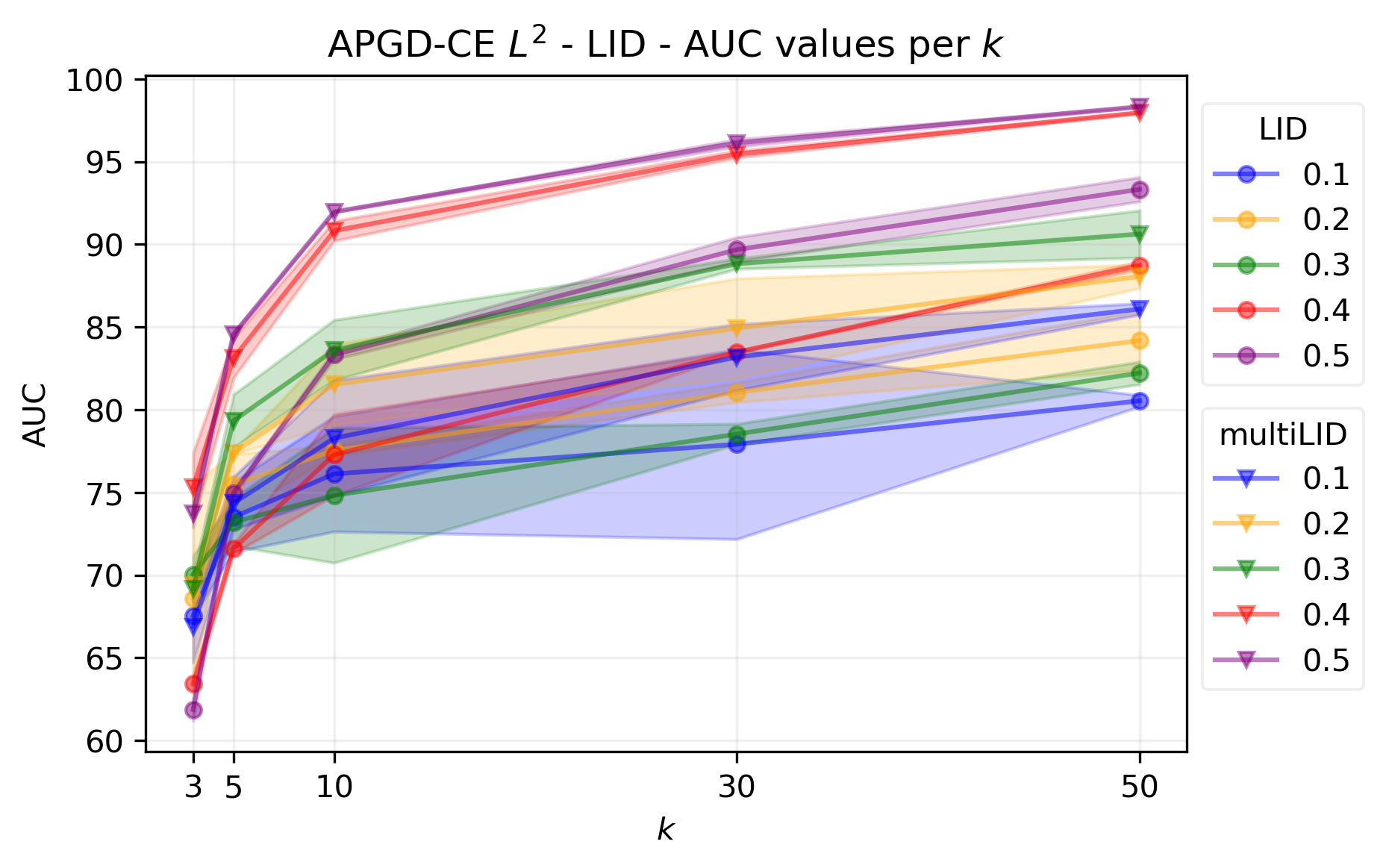}
      \caption{
      \corrected{
      The attack APGD-CE $L^{2}$ evaluated on different epsilons and neighbors.
      }
      }\label{fig:multil2eps}
    \end{subfigure}
    \caption{
    \corrected{
    Ablation study on CIFAR10 of LID and multiLID detection rates by using different $k$ on the APGD-CE ($L^2$, $L^{\infty}$) attack and different epsilon sizes.
    }
    } \label{fig:compeps}
    
\end{figure}

    We train the LID with the APGD-CE attack from the AutoAttack benchmark with different epsilons ($L^{\infty}$ and $L^{2}$). In \cref{fig:compeps}, we compare \ac{rf} and \ac{lr} on different norms. Random Forest succeeds on all epsilon sizes\footnote{ Perturbed images would round the adversarial changes to the next of 256 available bins in commonly used 8-bit per channel image encodings.} on both norms. On smaller perturbation sizes the \ac{lr} classifier AUC score falls. On the optimal perturbation size ($L^{\infty}: \epsilon=8/255$ and $L^{2}: \epsilon=0.5$) the \ac{lr} shows its best AUC scores. The \ac{rf} classifier gives us outstanding results over the \ac{lr}. Moreover, to save computation time, $k=3$ neighbors would be enough for high accuracy.





\FloatBarrier

\section{Conclusion}   \label{conclusion}

In this paper, we revisit the MLE estimate of the local intrinsic dimensionality which has been used in previous works on adversarial detection. An analysis of the extracted LID features and their theoretical properties allows us to redefine an LID-based feature using unfolded local growth rate estimates that are significantly more discriminative than the aggregated LID measure.\\ 
%

\noindent\textbf{Limitations.} While our method allows us to achieve almost perfect-to-perfect results in the considered test scenario and for the given datasets, we do not claim to have solved the actual problem. We use the evaluation setting as proposed in previous works (e.g.\cite{ma2018characterizing}) where each attack method is evaluated separately and with constant attack parameters. For deployment in real-world scenarios, the robustness of a detector under potential disguise mechanisms needs to be verified. An extended study on the transferability of our method from one attack to the other can be found in the supplementary material. It shows first promising results in this respect but also leaves room for further improvement.

%



\bibliographystyle{apalike}
{\small

}

\section*{\uppercase{Appendix}}

 
 \subsection*{A. Impact of the Number of Neighbors and Attack Strength $\epsilon$.}

We train LID and multiLID with the APGD-CE attack from the AutoAttack benchmark for different perturbation magnitudes, i.e.~using different epsilons ($L^{\infty}$ and $L^{2}$). On smaller perturbation sizes the \acf{lr} classifier AUC scores are dropping, which is to be expected. On the most commonly used perturbation sizes ($L^{\infty}: \epsilon=8/255$ and $L^{2}: \epsilon=0.5$) LID shows its best AUC scores. The multiLID classifier provides superior results over LID in all cases. Moreover, to save computation time for multiLID, $k=10$ neighbors would be enough for high-accuracy adversarial detection.

\subsection*{B. Attack Transferability} 

In this section, we evaluate the attack transferability of our models, for LID in \cref{tab:at1} and multiLID in \cref{tab:at1}. In the case of real-world applications, the attack methods might be unknown and thus it is a desired feature that a detector trained on one attack method performs well for a different attack. We evaluate in both directions. 
The \acf{rf} classifier shows significantly higher transferability on both LID and multiLID. The attack tuples (pgd $\leftrightarrow$ bim), (pgd $\leftrightarrow$ aa), (aa $\leftrightarrow$ bim), and (df $\leftrightarrow$ cw)  yield very high bidirectional attack transferability. However, the experiments also show that not all combinations can be transferred successfully, e.g. (fgsm $\leftrightarrow$ cw) in ImageNet. This leaves room for further research.

\begin{table*}[h]
\centering
\caption{Attack transfer LID. Rows with the target $\mu$ give the average transfer rates from one attack to all others. \ac{rf} shows higher accuracy (ACC) for the attack transfer.} \label{tab:at1}
\begin{adjustbox}{width=\textwidth}
\corrected{
\begin{tabular}{llrrrrrrrrrr}\hline \hline
    \multicolumn{12}{c}{\textbf{LID}} \\ \hline \hline
    \multicolumn{2}{c}{\multirow{2}{*}{\textbf{Attacks}}} & \multicolumn{4}{c}{\textbf{CIFAR10}} & \multicolumn{4}{c}{\textbf{CIFAR100}} & \multicolumn{2}{c}{\textbf{ImageNet}} \\ \cline{3-12} 
    \multicolumn{2}{c}{} & \multicolumn{2}{c}{\textbf{WRN 28-10}} & \multicolumn{2}{c}{\textbf{VGG16}} & \multicolumn{2}{c}{\textbf{WRN 28-10}} & \multicolumn{2}{c}{\textbf{VGG16}} & \multicolumn{2}{c}{\textbf{WRN 50-2}} \\ \hline
    \multicolumn{1}{c}{\textbf{from}} & \multicolumn{1}{c|}{\textbf{to}} & \multicolumn{1}{c}{\textbf{AUC}} & \multicolumn{1}{c|}{\textbf{ACC}} & \multicolumn{1}{c}{\textbf{AUC}} & \multicolumn{1}{c|}{\textbf{ACC}} & \multicolumn{1}{c}{\textbf{AUC}} & \multicolumn{1}{c|}{\textbf{ACC}} & \multicolumn{1}{c}{\textbf{AUC}} & \multicolumn{1}{c|}{\textbf{ACC}} & \multicolumn{1}{c}{\textbf{AUC}} & \multicolumn{1}{c}{\textbf{ACC}} \\ \hline\hline
    \multicolumn{12}{c}{\textbf{logistic regression}} \\ \hline
    \textbf{FGSM} & \textbf{BIM} & $62.3 \pm 102.5$ & $55.7 \pm 35.5$ & $62.9 \pm 118.6$ & $57.3 \pm 33.1$ & $66.8 \pm 191.5$ & $62.3 \pm 107.5$ & $79.2 \pm 0.8$ & $64.4 \pm 3.9$ & $76.3 \pm 0.2$ & $62.1 \pm 1.2$ \\
    \textbf{FGSM} & \textbf{PGD} & $78.3 \pm 65.7$ & $61.2 \pm 208.8$ & $86.9 \pm 0.4$ & $77.7 \pm 0.5$ & $95.6 \pm 1.0$ & $87.2 \pm 1.8$ & $66.1 \pm 273.0$ & $61.1 \pm 217.0$ & $60.6 \pm 265.3$ & $58.3 \pm 137.2$ \\
    \textbf{FGSM} & \textbf{AA} & $78.9 \pm 68.6$ & $60.8 \pm 229.0$ & $86.9 \pm 0.3$ & $79.5 \pm 0.5$ & $96.5 \pm 0.2$ & $88.5 \pm 0.7$ & $66.6 \pm 321.8$ & $61.7 \pm 268.3$ & $62.0 \pm 319.2$ & $59.4 \pm 176.1$ \\
    \textbf{FGSM} & \textbf{DF} & $79.4 \pm 45.2$ & $60.4 \pm 191.4$ & $86.4 \pm 0.0$ & $74.0 \pm 1.3$ & $87.4 \pm 2.2$ & $74.3 \pm 2.2$ & $66.5 \pm 284.8$ & $60.8 \pm 204.7$ & $61.7 \pm 280.9$ & $58.4 \pm 136.7$ \\
    \textbf{FGSM} & \textbf{CW} & $87.8 \pm 0.9$ & $79.4 \pm 0.0$ & $67.5 \pm 70.9$ & $55.0 \pm 50.0$ & $67.2 \pm 99.5$ & $54.8 \pm 46.1$ & $72.5 \pm 227.6$ & $61.7 \pm 270.7$ & $92.8 \pm 0.3$ & $82.9 \pm 0.4$ \\ 
    \hline 
    \textbf{FGSM} & \textbf{$\mu$} & $69.5 \pm 82.7$ & $60.4 \pm 36.3$ & $77.5 \pm 121.1$ & $69.1 \pm 113.1$ & $78.2 \pm 142.0$ & $70.0 \pm 134.9$ & $76.3 \pm 122.6$ & $65.6 \pm 107.3$ & $77.5 \pm 79.8$ & $66.8 \pm 73.4$\\ 
    \hline
    \textbf{BIM} & \textbf{FGSM} & $85.5 \pm 0.8$ & $77.8 \pm 0.2$ & $66.3 \pm 43.3$ & $54.6 \pm 41.7$ & $66.5 \pm 72.4$ & $54.7 \pm 43.6$ & $72.0 \pm 192.2$ & $60.8 \pm 233.3$ & $94.0 \pm 0.2$ & $86.6 \pm 1.0$ \\
    \textbf{BIM} & \textbf{PGD} & $81.0 \pm 0.6$ & $74.0 \pm 0.4$ & $78.8 \pm 0.9$ & $72.0 \pm 1.0$ & $92.0 \pm 1.7$ & $84.6 \pm 2.6$ & $78.8 \pm 0.1$ & $69.2 \pm 0.4$ & $75.9 \pm 0.4$ & $66.8 \pm 0.1$ \\
    \textbf{BIM} & \textbf{AA} & $81.3 \pm 0.6$ & $72.9 \pm 1.4$ & $85.0 \pm 0.2$ & $75.5 \pm 0.5$ & $95.6 \pm 0.4$ & $87.9 \pm 1.1$ & $80.5 \pm 0.0$ & $71.6 \pm 0.1$ & $78.3 \pm 0.4$ & $69.0 \pm 2.0$ \\
    \textbf{BIM} & \textbf{DF} & $81.6 \pm 0.1$ & $73.5 \pm 0.6$ & $87.6 \pm 0.8$ & $79.5 \pm 0.5$ & $95.5 \pm 0.3$ & $87.3 \pm 1.6$ & $79.2 \pm 0.9$ & $71.4 \pm 0.5$ & $77.2 \pm 0.8$ & $68.7 \pm 0.3$ \\
    \textbf{BIM} & \textbf{CW} & $82.7 \pm 0.1$ & $71.2 \pm 0.4$ & $86.7 \pm 0.4$ & $75.4 \pm 0.4$ & $84.1 \pm 0.3$ & $71.8 \pm 0.5$ & $80.7 \pm 1.4$ & $67.1 \pm 0.1$ & $78.2 \pm 1.1$ & $63.2 \pm 0.3$ \\ 
    \hline 
    \textbf{BIM} & \textbf{$\mu$} & $76.8 \pm 61.8$ & $66.9 \pm 63.9$ & $81.3 \pm 0.8$ & $73.3 \pm 0.9$ & $84.1 \pm 0.3$ & $75.4 \pm 1.0$ & $84.2 \pm 0.6$ & $76.1 \pm 0.7$ & $82.5 \pm 0.7$ & $69.7 \pm 0.4$\\ 
    \hline

\textbf{PGD} & \textbf{FGSM} & $80.7 \pm 0.2$ & $72.6 \pm 0.4$ & $83.6 \pm 0.7$ & $76.1 \pm 2.6$ & $80.5 \pm 0.4$ & $73.8 \pm 0.8$ & $93.8 \pm 0.1$ & $85.3 \pm 0.1$ & $82.9 \pm 0.5$ & $74.5 \pm 0.1$ \\
\textbf{PGD} & \textbf{BIM} & $79.8 \pm 0.1$ & $72.6 \pm 0.5$ & $84.2 \pm 0.4$ & $77.4 \pm 1.2$ & $81.1 \pm 0.2$ & $74.6 \pm 1.8$ & $94.0 \pm 0.1$ & $85.2 \pm 0.2$ & $86.3 \pm 0.2$ & $77.5 \pm 0.2$ \\
\textbf{PGD} & \textbf{AA} & $90.1 \pm 10.2$ & $78.8 \pm 10.5$ & $91.0 \pm 4.2$ & $80.3 \pm 15.4$ & $94.8 \pm 4.8$ & $87.6 \pm 9.1$ & $54.5 \pm 0.2$ & $52.1 \pm 0.1$ & $52.1 \pm 0.1$ & $50.7 \pm 0.3$ \\
\textbf{PGD} & \textbf{DF} & $94.2 \pm 2.9$ & $84.1 \pm 8.8$ & $97.7 \pm 0.1$ & $91.9 \pm 1.1$ & $99.3 \pm 0.0$ & $94.8 \pm 0.2$ & $52.4 \pm 1.1$ & $49.2 \pm 0.3$ & $50.9 \pm 0.2$ & $48.9 \pm 0.1$ \\
\textbf{PGD} & \textbf{CW} & $93.6 \pm 5.8$ & $77.2 \pm 48.2$ & $95.2 \pm 0.9$ & $87.3 \pm 1.8$ & $99.3 \pm 0.0$ & $94.9 \pm 0.2$ & $52.1 \pm 1.0$ & $49.7 \pm 0.1$ & $51.2 \pm 0.2$ & $49.7 \pm 0.3$ \\ 
\hline 
\textbf{PGD} & \textbf{$\mu$} & $84.3 \pm 0.4$ & $76.5 \pm 0.8$ & $85.1 \pm 0.2$ & $77.5 \pm 0.8$ & $76.5 \pm 3.9$ & $69.9 \pm 7.1$ & $78.9 \pm 0.9$ & $73.8 \pm 2.1$ & $78.3 \pm 1.6$ & $71.7 \pm 10.1$\\ 
\hline

\textbf{AA} & \textbf{FGSM} & $93.4 \pm 2.7$ & $71.2 \pm 15.6$ & $94.4 \pm 1.2$ & $80.8 \pm 8.5$ & $97.0 \pm 0.1$ & $87.4 \pm 2.1$ & $52.2 \pm 0.9$ & $50.0 \pm 0.0$ & $50.6 \pm 0.0$ & $49.9 \pm 0.1$ \\
\textbf{AA} & \textbf{BIM} & $93.3 \pm 1.2$ & $78.5 \pm 0.8$ & $40.9 \pm 114.9$ & $48.2 \pm 40.9$ & $29.2 \pm 128.7$ & $41.1 \pm 42.8$ & $20.8 \pm 63.1$ & $37.1 \pm 12.8$ & $52.5 \pm 0.0$ & $51.5 \pm 0.2$ \\
\textbf{AA} & \textbf{PGD} & $87.7 \pm 2.5$ & $72.8 \pm 4.2$ & $39.3 \pm 3.5$ & $42.9 \pm 3.4$ & $30.2 \pm 5.3$ & $38.3 \pm 9.7$ & $20.0 \pm 3.6$ & $33.9 \pm 4.4$ & $54.1 \pm 0.6$ & $52.6 \pm 0.2$ \\
\textbf{AA} & \textbf{DF} & $77.7 \pm 0.9$ & $70.3 \pm 0.0$ & $79.4 \pm 0.5$ & $72.9 \pm 0.6$ & $83.5 \pm 0.4$ & $76.4 \pm 0.2$ & $53.3 \pm 0.2$ & $50.5 \pm 0.2$ & $54.7 \pm 1.1$ & $52.0 \pm 0.4$ \\
\textbf{AA} & \textbf{CW} & $70.6 \pm 0.3$ & $64.2 \pm 1.0$ & $83.0 \pm 2.4$ & $75.9 \pm 3.1$ & $86.3 \pm 0.2$ & $79.1 \pm 0.4$ & $52.8 \pm 0.7$ & $50.2 \pm 0.6$ & $56.8 \pm 0.4$ & $51.3 \pm 1.5$ \\ 
\hline 
\textbf{AA} & \textbf{$\mu$} & $77.5 \pm 1.0$ & $67.9 \pm 5.3$ & $47.3 \pm 61.6$ & $51.3 \pm 19.5$ & $46.2 \pm 3.1$ & $48.1 \pm 4.4$ & $69.7 \pm 0.6$ & $64.4 \pm 0.3$ & $69.9 \pm 0.8$ & $64.1 \pm 1.3$\\ 
\hline

\textbf{DF} & \textbf{FGSM} & $69.5 \pm 0.3$ & $64.0 \pm 0.7$ & $80.3 \pm 1.3$ & $72.6 \pm 4.0$ & $83.7 \pm 1.0$ & $77.4 \pm 0.2$ & $52.5 \pm 0.5$ & $50.6 \pm 1.6$ & $58.8 \pm 0.7$ & $51.8 \pm 1.6$ \\
\textbf{DF} & \textbf{BIM} & $71.9 \pm 0.8$ & $62.2 \pm 0.6$ & $80.3 \pm 4.8$ & $71.9 \pm 4.5$ & $80.6 \pm 1.8$ & $71.6 \pm 2.7$ & $53.0 \pm 0.2$ & $49.6 \pm 1.1$ & $54.3 \pm 0.1$ & $49.6 \pm 1.2$ \\
\textbf{DF} & \textbf{PGD} & $68.4 \pm 2.4$ & $64.1 \pm 3.4$ & $72.5 \pm 1.0$ & $65.4 \pm 0.1$ & $74.4 \pm 0.6$ & $68.6 \pm 0.1$ & $79.3 \pm 12.4$ & $70.0 \pm 11.2$ & $55.6 \pm 0.2$ & $54.1 \pm 0.7$ \\
\textbf{DF} & \textbf{AA} & $61.9 \pm 2.4$ & $58.7 \pm 2.3$ & $73.7 \pm 11.5$ & $66.1 \pm 2.8$ & $82.9 \pm 7.1$ & $71.1 \pm 2.7$ & $77.5 \pm 10.9$ & $65.8 \pm 1.8$ & $52.5 \pm 0.3$ & $52.0 \pm 0.3$ \\
\textbf{DF} & \textbf{CW} & $70.1 \pm 10.0$ & $64.5 \pm 5.7$ & $78.4 \pm 2.4$ & $71.5 \pm 1.8$ & $94.2 \pm 5.8$ & $77.5 \pm 6.0$ & $51.5 \pm 0.1$ & $51.1 \pm 0.9$ & $51.3 \pm 0.0$ & $50.9 \pm 0.0$ \\ 
\hline 
\textbf{DF} & \textbf{$\mu$} & $69.0 \pm 0.8$ & $63.3 \pm 1.6$ & $68.0 \pm 1.5$ & $61.0 \pm 2.0$ & $70.0 \pm 3.3$ & $64.4 \pm 3.1$ & $69.7 \pm 6.5$ & $62.7 \pm 2.0$ & $69.1 \pm 3.7$ & $63.1 \pm 2.9$\\ 
\hline
\textbf{CW} & \textbf{FGSM} & $56.5 \pm 1.1$ & $51.3 \pm 0.4$ & $93.8 \pm 1.3$ & $87.0 \pm 2.0$ & $99.9 \pm 0.0$ & $95.8 \pm 0.2$ & $54.2 \pm 0.1$ & $51.6 \pm 0.1$ & $54.1 \pm 0.1$ & $50.9 \pm 0.1$ \\
\textbf{CW} & \textbf{BIM} & $59.0 \pm 2.3$ & $51.2 \pm 0.3$ & $91.8 \pm 2.7$ & $82.2 \pm 2.6$ & $100.0 \pm 0.0$ & $97.0 \pm 0.2$ & $54.1 \pm 0.0$ & $51.0 \pm 0.1$ & $53.9 \pm 0.0$ & $51.3 \pm 0.1$ \\
\textbf{CW} & \textbf{PGD} & $52.7 \pm 1.0$ & $49.8 \pm 0.2$ & $89.5 \pm 1.8$ & $62.9 \pm 3.4$ & $88.0 \pm 1.8$ & $63.6 \pm 1.2$ & $53.4 \pm 0.0$ & $50.2 \pm 0.0$ & $53.1 \pm 0.0$ & $50.0 \pm 0.0$ \\
\textbf{CW} & \textbf{AA} & $57.4 \pm 3.1$ & $54.8 \pm 1.1$ & $92.6 \pm 0.6$ & $77.6 \pm 0.7$ & $94.9 \pm 0.0$ & $78.2 \pm 1.0$ & $99.8 \pm 0.0$ & $79.8 \pm 0.4$ & $54.2 \pm 0.0$ & $52.5 \pm 0.1$ \\
\textbf{CW} & \textbf{DF} & $56.2 \pm 0.0$ & $54.7 \pm 0.9$ & $92.5 \pm 2.2$ & $75.6 \pm 2.0$ & $94.5 \pm 0.9$ & $77.4 \pm 2.6$ & $99.8 \pm 0.0$ & $79.1 \pm 2.7$ & $54.4 \pm 0.1$ & $53.0 \pm 0.5$ \\ 
\hline 
\textbf{CW} & \textbf{$\mu$} & $71.7 \pm 0.5$ & $67.3 \pm 0.6$ & $71.8 \pm 1.0$ & $66.5 \pm 0.7$ & $67.3 \pm 0.9$ & $55.3 \pm 1.0$ & $79.8 \pm 0.8$ & $68.6 \pm 0.7$ & $79.5 \pm 0.7$ & $68.0 \pm 1.7$\\ 
\hline
    \multicolumn{12}{c}{\textbf{random forest}} \\ \hline
\textbf{FGSM} & \textbf{BIM} & $62.3 \pm 166.8$ & $57.9 \pm 79.1$ & $61.0 \pm 220.7$ & $55.7 \pm 100.7$ & $68.5 \pm 243.0$ & $62.4 \pm 198.1$ & $79.6 \pm 0.2$ & $66.5 \pm 2.0$ & $75.7 \pm 0.6$ & $64.9 \pm 7.3$ \\
\textbf{FGSM} & \textbf{PGD} & $54.4 \pm 498.7$ & $54.1 \pm 238.9$ & $90.0 \pm 1.1$ & $82.3 \pm 3.9$ & $94.8 \pm 1.4$ & $88.0 \pm 2.0$ & $64.4 \pm 133.6$ & $59.8 \pm 61.7$ & $59.1 \pm 130.7$ & $56.0 \pm 49.3$ \\
\textbf{FGSM} & \textbf{AA} & $55.6 \pm 535.7$ & $55.5 \pm 355.3$ & $87.3 \pm 0.1$ & $79.6 \pm 0.2$ & $95.6 \pm 0.9$ & $89.2 \pm 1.0$ & $66.3 \pm 216.2$ & $61.1 \pm 141.7$ & $61.1 \pm 224.0$ & $57.4 \pm 115.8$ \\
\textbf{FGSM} & \textbf{DF} & $58.3 \pm 499.3$ & $57.9 \pm 234.0$ & $86.3 \pm 1.4$ & $73.2 \pm 0.4$ & $88.7 \pm 5.8$ & $77.7 \pm 5.2$ & $69.3 \pm 199.5$ & $62.0 \pm 104.3$ & $62.8 \pm 248.7$ & $57.0 \pm 91.6$ \\
\textbf{FGSM} & \textbf{CW} & $86.5 \pm 0.9$ & $78.3 \pm 0.5$ & $62.5 \pm 102.5$ & $54.7 \pm 44.2$ & $66.6 \pm 100.0$ & $56.2 \pm 76.1$ & $71.5 \pm 276.0$ & $63.3 \pm 352.0$ & $93.4 \pm 0.3$ & $87.3 \pm 0.1$ \\ 
\hline 
\textbf{FGSM} & \textbf{$\mu$} & $69.4 \pm 126.3$ & $61.5 \pm 77.4$ & $72.5 \pm 153.1$ & $68.0 \pm 71.2$ & $73.2 \pm 195.4$ & $68.5 \pm 122.8$ & $73.1 \pm 191.0$ & $65.6 \pm 87.1$ & $76.1 \pm 96.0$ & $68.0 \pm 94.6$\\ 
\hline

\textbf{BIM} & \textbf{FGSM} & $84.5 \pm 0.4$ & $77.3 \pm 0.4$ & $53.9 \pm 231.5$ & $55.5 \pm 61.6$ & $59.2 \pm 230.5$ & $56.8 \pm 93.4$ & $66.1 \pm 418.1$ & $62.9 \pm 335.4$ & $94.6 \pm 0.1$ & $89.2 \pm 0.8$ \\
\textbf{BIM} & \textbf{PGD} & $79.4 \pm 0.1$ & $70.5 \pm 0.5$ & $75.6 \pm 0.6$ & $67.6 \pm 0.4$ & $88.2 \pm 3.5$ & $79.5 \pm 0.5$ & $76.1 \pm 0.0$ & $67.3 \pm 0.9$ & $73.4 \pm 0.2$ & $65.5 \pm 0.1$ \\
\textbf{BIM} & \textbf{AA} & $75.4 \pm 0.5$ & $64.9 \pm 0.1$ & $90.3 \pm 0.5$ & $82.0 \pm 0.7$ & $97.6 \pm 0.2$ & $91.5 \pm 0.1$ & $77.1 \pm 2.4$ & $66.8 \pm 5.1$ & $74.8 \pm 1.3$ & $62.8 \pm 1.4$ \\
\textbf{BIM} & \textbf{DF} & $74.9 \pm 0.4$ & $66.2 \pm 0.1$ & $91.9 \pm 1.2$ & $84.3 \pm 1.2$ & $97.4 \pm 0.1$ & $89.8 \pm 0.4$ & $75.0 \pm 3.1$ & $65.5 \pm 3.7$ & $73.5 \pm 0.2$ & $63.3 \pm 0.4$ \\
\textbf{BIM} & \textbf{CW} & $79.3 \pm 0.1$ & $64.5 \pm 0.7$ & $90.9 \pm 1.2$ & $81.5 \pm 0.3$ & $88.4 \pm 0.3$ & $78.3 \pm 0.3$ & $80.2 \pm 1.0$ & $65.3 \pm 0.8$ & $77.7 \pm 0.2$ & $60.7 \pm 0.9$ \\ 
\hline 
\textbf{BIM} & \textbf{$\mu$} & $71.7 \pm 176.1$ & $68.3 \pm 98.3$ & $78.5 \pm 0.9$ & $70.1 \pm 0.5$ & $83.0 \pm 1.0$ & $73.6 \pm 1.5$ & $82.5 \pm 1.0$ & $73.8 \pm 1.2$ & $83.3 \pm 0.6$ & $70.0 \pm 0.6$\\ 
\hline

\textbf{PGD} & \textbf{FGSM} & $78.8 \pm 0.7$ & $71.7 \pm 0.8$ & $81.9 \pm 0.7$ & $75.4 \pm 0.8$ & $78.9 \pm 0.7$ & $72.8 \pm 2.2$ & $91.5 \pm 0.2$ & $82.6 \pm 0.3$ & $83.3 \pm 0.7$ & $76.8 \pm 0.1$ \\
\textbf{PGD} & \textbf{BIM} & $76.4 \pm 0.8$ & $71.8 \pm 0.7$ & $79.8 \pm 0.1$ & $74.2 \pm 0.7$ & $77.5 \pm 0.0$ & $72.2 \pm 0.2$ & $88.1 \pm 3.6$ & $82.1 \pm 1.6$ & $85.5 \pm 0.0$ & $79.5 \pm 0.2$ \\
\textbf{PGD} & \textbf{AA} & $80.7 \pm 11.2$ & $68.1 \pm 1.2$ & $79.0 \pm 11.8$ & $67.1 \pm 3.4$ & $83.3 \pm 3.7$ & $70.4 \pm 0.4$ & $53.4 \pm 0.1$ & $52.6 \pm 0.0$ & $51.7 \pm 0.2$ & $51.2 \pm 0.1$ \\
\textbf{PGD} & \textbf{DF} & $89.7 \pm 0.8$ & $73.7 \pm 7.4$ & $97.7 \pm 0.0$ & $92.3 \pm 0.4$ & $99.1 \pm 0.0$ & $94.6 \pm 0.3$ & $52.1 \pm 2.5$ & $48.7 \pm 0.7$ & $50.5 \pm 0.0$ & $48.8 \pm 0.1$ \\
\textbf{PGD} & \textbf{CW} & $90.5 \pm 2.7$ & $67.9 \pm 38.9$ & $95.2 \pm 0.4$ & $88.7 \pm 0.3$ & $99.2 \pm 0.0$ & $95.2 \pm 0.6$ & $49.9 \pm 0.8$ & $49.0 \pm 0.2$ & $49.0 \pm 0.2$ & $48.7 \pm 0.1$ \\ 
\hline 
\textbf{PGD} & \textbf{$\mu$} & $82.9 \pm 0.6$ & $75.9 \pm 0.8$ & $81.5 \pm 0.9$ & $75.9 \pm 0.7$ & $69.6 \pm 5.4$ & $61.9 \pm 1.0$ & $77.8 \pm 0.7$ & $71.6 \pm 1.8$ & $76.8 \pm 0.8$ & $69.9 \pm 8.0$\\ 
\hline

\textbf{AA} & \textbf{FGSM} & $89.0 \pm 12.1$ & $58.6 \pm 11.9$ & $94.8 \pm 0.5$ & $86.3 \pm 0.4$ & $97.5 \pm 0.1$ & $91.6 \pm 0.2$ & $49.7 \pm 0.5$ & $49.2 \pm 0.0$ & $48.8 \pm 0.6$ & $49.3 \pm 0.1$ \\
\textbf{AA} & \textbf{BIM} & $70.3 \pm 9.3$ & $64.0 \pm 7.3$ & $20.1 \pm 7.3$ & $34.8 \pm 0.1$ & $14.9 \pm 0.5$ & $30.6 \pm 4.1$ & $10.0 \pm 3.5$ & $29.1 \pm 2.1$ & $58.7 \pm 0.7$ & $56.3 \pm 1.3$ \\
\textbf{AA} & \textbf{PGD} & $59.4 \pm 5.8$ & $57.3 \pm 7.2$ & $22.9 \pm 16.7$ & $32.9 \pm 4.2$ & $19.1 \pm 17.7$ & $29.9 \pm 4.6$ & $14.2 \pm 25.7$ & $27.5 \pm 2.7$ & $60.7 \pm 1.9$ & $58.0 \pm 2.2$ \\
\textbf{AA} & \textbf{DF} & $73.0 \pm 0.7$ & $66.8 \pm 0.2$ & $77.7 \pm 0.8$ & $71.4 \pm 0.7$ & $78.3 \pm 0.9$ & $71.3 \pm 1.3$ & $55.7 \pm 0.2$ & $53.5 \pm 0.2$ & $56.6 \pm 0.3$ & $54.7 \pm 1.5$ \\
\textbf{AA} & \textbf{CW} & $70.7 \pm 1.9$ & $63.0 \pm 3.9$ & $85.3 \pm 1.8$ & $78.5 \pm 3.7$ & $89.4 \pm 0.6$ & $80.0 \pm 0.7$ & $53.1 \pm 0.0$ & $51.4 \pm 1.2$ & $57.9 \pm 0.4$ & $53.9 \pm 0.3$ \\ 
\hline 
\textbf{AA} & \textbf{$\mu$} & $76.0 \pm 2.8$ & $67.0 \pm 2.5$ & $34.8 \pm 4.3$ & $42.9 \pm 3.0$ & $35.3 \pm 13.6$ & $41.1 \pm 4.2$ & $68.3 \pm 0.6$ & $63.5 \pm 0.8$ & $71.3 \pm 1.0$ & $65.4 \pm 2.0$\\ 
\hline

\textbf{DF} & \textbf{FGSM} & $70.5 \pm 1.3$ & $59.8 \pm 0.4$ & $80.5 \pm 3.8$ & $72.4 \pm 3.8$ & $88.4 \pm 0.8$ & $80.2 \pm 0.8$ & $53.7 \pm 2.0$ & $52.1 \pm 1.2$ & $57.8 \pm 0.2$ & $54.2 \pm 0.9$ \\
\textbf{DF} & \textbf{BIM} & $70.3 \pm 0.9$ & $60.5 \pm 0.2$ & $82.4 \pm 5.2$ & $73.6 \pm 6.0$ & $85.7 \pm 1.6$ & $77.8 \pm 2.7$ & $52.8 \pm 0.0$ & $50.1 \pm 0.7$ & $56.8 \pm 0.3$ & $53.0 \pm 0.9$ \\
\textbf{DF} & \textbf{PGD} & $58.8 \pm 0.6$ & $56.1 \pm 0.4$ & $51.4 \pm 2.1$ & $49.9 \pm 3.5$ & $56.0 \pm 2.4$ & $54.2 \pm 2.5$ & $46.6 \pm 3.7$ & $47.1 \pm 0.7$ & $60.9 \pm 2.0$ & $57.4 \pm 0.7$ \\
\textbf{DF} & \textbf{AA} & $58.4 \pm 2.4$ & $56.7 \pm 1.0$ & $62.2 \pm 3.4$ & $60.1 \pm 1.3$ & $67.8 \pm 3.0$ & $63.4 \pm 1.9$ & $56.2 \pm 7.0$ & $55.0 \pm 9.5$ & $59.0 \pm 0.1$ & $56.5 \pm 0.1$ \\
\textbf{DF} & \textbf{CW} & $59.4 \pm 5.9$ & $55.9 \pm 2.0$ & $62.1 \pm 0.3$ & $58.5 \pm 6.0$ & $56.3 \pm 7.1$ & $52.9 \pm 5.7$ & $50.6 \pm 0.1$ & $50.4 \pm 1.2$ & $50.5 \pm 0.1$ & $50.5 \pm 0.1$ \\ 
\hline 
\textbf{DF} & \textbf{$\mu$} & $70.2 \pm 1.6$ & $63.7 \pm 1.4$ & $69.6 \pm 1.6$ & $63.0 \pm 2.1$ & $54.7 \pm 2.2$ & $52.9 \pm 1.6$ & $60.7 \pm 3.2$ & $58.3 \pm 2.8$ & $55.8 \pm 2.7$ & $53.6 \pm 3.0$\\ 
\hline

\textbf{CW} & \textbf{FGSM} & $54.7 \pm 3.1$ & $52.8 \pm 0.5$ & $92.9 \pm 1.2$ & $85.8 \pm 1.4$ & $99.4 \pm 0.1$ & $94.6 \pm 0.1$ & $53.7 \pm 0.1$ & $51.4 \pm 0.1$ & $53.0 \pm 0.0$ & $50.8 \pm 0.3$ \\
\textbf{CW} & \textbf{BIM} & $55.3 \pm 5.2$ & $52.0 \pm 0.8$ & $91.3 \pm 1.7$ & $82.7 \pm 3.9$ & $99.7 \pm 0.0$ & $96.0 \pm 0.1$ & $53.6 \pm 0.2$ & $51.0 \pm 0.0$ & $53.1 \pm 0.3$ & $51.4 \pm 0.2$ \\
\textbf{CW} & \textbf{PGD} & $48.3 \pm 1.9$ & $49.8 \pm 0.0$ & $85.5 \pm 4.0$ & $64.9 \pm 5.3$ & $88.5 \pm 7.5$ & $73.0 \pm 1.4$ & $51.3 \pm 0.2$ & $50.3 \pm 0.0$ & $52.0 \pm 1.9$ & $50.2 \pm 0.0$ \\
\textbf{CW} & \textbf{AA} & $54.1 \pm 2.4$ & $53.0 \pm 1.0$ & $71.7 \pm 4.9$ & $65.7 \pm 2.8$ & $73.5 \pm 27.9$ & $66.0 \pm 6.0$ & $67.6 \pm 140.4$ & $62.3 \pm 90.2$ & $53.7 \pm 1.8$ & $52.7 \pm 0.3$ \\
\textbf{CW} & \textbf{DF} & $52.3 \pm 1.5$ & $51.9 \pm 1.3$ & $72.1 \pm 0.3$ & $65.5 \pm 1.8$ & $79.6 \pm 4.6$ & $68.6 \pm 1.0$ & $80.8 \pm 124.5$ & $68.3 \pm 44.4$ & $53.7 \pm 0.7$ & $53.0 \pm 1.1$ \\ 
\hline 
\textbf{CW} & \textbf{$\mu$} & $70.7 \pm 0.9$ & $67.1 \pm 0.5$ & $70.6 \pm 1.5$ & $66.6 \pm 1.0$ & $65.1 \pm 3.1$ & $57.6 \pm 1.4$ & $64.1 \pm 35.5$ & $59.9 \pm 20.1$ & $67.7 \pm 26.3$ & $61.4 \pm 9.9$\\  
\hline
\bottomrule[1pt]
\end{tabular}
}
\end{adjustbox}
\end{table*}

\begin{table*}[h]
\centering
\caption{Attack transfer multiLID. Rows with the target $\mu$ give the average transfer rates from one attack to all others. The full multiLID with \ac{rf} shows significantly better accuracy (ACC) for the attack transfer.} \label{tab:at2}
\begin{adjustbox}{width=\textwidth}
\corrected{
\begin{tabular}{llrrrrrrrrrr}
\hline \hline
\multicolumn{12}{c}{\textbf{multiLID}} \\ \hline \hline
\multicolumn{2}{c}{\multirow{2}{*}{\textbf{Attacks}}} & \multicolumn{4}{c}{\textbf{CIFAR10}} & \multicolumn{4}{c}{\textbf{CIFAR100}} & \multicolumn{2}{c}{\textbf{ImageNet}} \\ \cline{3-12} 
\multicolumn{2}{c}{} & \multicolumn{2}{c}{\textbf{WRN 28-10}} & \multicolumn{2}{c}{\textbf{VGG16}} & \multicolumn{2}{c}{\textbf{WRN 28-10}} & \multicolumn{2}{c}{\textbf{VGG16}} & \multicolumn{2}{c}{\textbf{WRN 50-2}} \\ \hline
\multicolumn{1}{c}{\textbf{from}} & \multicolumn{1}{c|}{\textbf{to}} &\multicolumn{1}{c}{\textbf{AUC}} & \multicolumn{1}{c|}{\textbf{ACC}} & \multicolumn{1}{c}{\textbf{AUC}} & \multicolumn{1}{c|}{\textbf{ACC}} & \multicolumn{1}{c}{\textbf{AUC}} & \multicolumn{1}{c|}{\textbf{ACC}} & \multicolumn{1}{c}{\textbf{AUC}} & \multicolumn{1}{c|}{\textbf{ACC}} & \multicolumn{1}{c}{\textbf{AUC}} & \multicolumn{1}{c}{\textbf{ACC}} \\ \hline \hline
\multicolumn{12}{c}{\textbf{logistic regression}} \\ \hline 
\textbf{FGSM} & \textbf{BIM} & $74.9 \pm 62.2$ & $59.5 \pm 164.0$ & $90.4 \pm 1.6$ & $76.5 \pm 33.9$ & $91.0 \pm 28.8$ & $76.1 \pm 249.5$ & $83.0 \pm 4.8$ & $71.7 \pm 10.6$ & $79.1 \pm 1.5$ & $68.0 \pm 5.0$ \\
\textbf{FGSM} & \textbf{PGD} & $42.4 \pm 1244.3$ & $61.3 \pm 292.9$ & $45.2 \pm 1322.9$ & $64.1 \pm 373.6$ & $69.2 \pm 1808.0$ & $78.7 \pm 410.9$ & $32.5 \pm 1507.4$ & $58.9 \pm 248.6$ & $30.3 \pm 1279.1$ & $55.3 \pm 197.6$ \\
\textbf{FGSM} & \textbf{AA} & $93.7 \pm 2.5$ & $86.0 \pm 4.5$ & $50.2 \pm 976.9$ & $63.3 \pm 334.6$ & $70.0 \pm 1672.5$ & $79.7 \pm 438.1$ & $88.0 \pm 0.5$ & $80.0 \pm 0.0$ & $83.7 \pm 4.9$ & $76.1 \pm 0.0$ \\
\textbf{FGSM} & \textbf{DF} & $69.1 \pm 1144.7$ & $72.9 \pm 286.9$ & $70.7 \pm 959.5$ & $71.3 \pm 219.2$ & $70.6 \pm 1203.8$ & $75.0 \pm 311.5$ & $61.5 \pm 1432.3$ & $68.9 \pm 222.8$ & $57.9 \pm 1367.7$ & $65.1 \pm 160.8$ \\
\textbf{FGSM} & \textbf{CW} & $87.0 \pm 1.9$ & $79.6 \pm 0.9$ & $57.0 \pm 587.2$ & $54.7 \pm 43.6$ & $87.1 \pm 4.5$ & $69.2 \pm 10.8$ & $66.4 \pm 1584.1$ & $77.6 \pm 383.0$ & $97.1 \pm 0.7$ & $93.7 \pm 0.3$ \\ 
\hline 
\textbf{FGSM} & \textbf{$\mu$} & $83.7 \pm 19.8$ & $70.4 \pm 92.6$ & $43.9 \pm 1432.3$ & $63.6 \pm 304.7$ & $77.1 \pm 531.5$ & $77.0 \pm 155.5$ & $66.0 \pm 1221.6$ & $70.7 \pm 240.2$ & $78.9 \pm 435.7$ & $75.0 \pm 87.7$\\ 
\hline

\textbf{BIM} & \textbf{FGSM} & $86.0 \pm 2.2$ & $77.8 \pm 3.8$ & $48.1 \pm 554.0$ & $52.2 \pm 9.4$ & $86.1 \pm 1.5$ & $64.0 \pm 0.5$ & $66.6 \pm 1507.0$ & $76.8 \pm 360.1$ & $97.6 \pm 0.1$ & $93.7 \pm 0.3$ \\
\textbf{BIM} & \textbf{PGD} & $78.3 \pm 0.6$ & $68.8 \pm 0.2$ & $75.5 \pm 1.1$ & $66.8 \pm 1.7$ & $89.5 \pm 0.6$ & $82.0 \pm 0.4$ & $78.1 \pm 1.6$ & $68.9 \pm 0.6$ & $78.3 \pm 1.1$ & $68.0 \pm 2.9$ \\
\textbf{BIM} & \textbf{AA} & $81.0 \pm 0.6$ & $71.8 \pm 0.7$ & $89.9 \pm 0.1$ & $82.1 \pm 2.3$ & $98.1 \pm 0.7$ & $91.2 \pm 0.7$ & $70.6 \pm 4.4$ & $63.7 \pm 0.9$ & $70.7 \pm 7.2$ & $63.0 \pm 11.5$ \\
\textbf{BIM} & \textbf{DF} & $82.5 \pm 2.7$ & $73.9 \pm 3.0$ & $92.6 \pm 1.4$ & $84.8 \pm 0.7$ & $97.8 \pm 0.6$ & $89.7 \pm 2.3$ & $69.4 \pm 0.3$ & $63.9 \pm 1.1$ & $69.4 \pm 1.4$ & $62.1 \pm 5.2$ \\
\textbf{BIM} & \textbf{CW} & $83.6 \pm 1.8$ & $70.6 \pm 1.3$ & $91.8 \pm 0.8$ & $85.6 \pm 0.9$ & $88.7 \pm 0.7$ & $81.0 \pm 2.5$ & $74.6 \pm 0.2$ & $63.5 \pm 1.1$ & $74.6 \pm 1.8$ & $61.9 \pm 2.7$ \\ 
\hline 
\textbf{BIM} & \textbf{$\mu$} & $76.9 \pm 412.9$ & $72.9 \pm 74.8$ & $80.0 \pm 1.0$ & $70.9 \pm 1.2$ & $82.1 \pm 2.6$ & $74.4 \pm 3.2$ & $82.4 \pm 1.3$ & $74.9 \pm 2.5$ & $82.7 \pm 1.1$ & $72.5 \pm 1.7$\\ 
\hline

\textbf{PGD} & \textbf{FGSM} & $81.9 \pm 0.3$ & $73.4 \pm 0.7$ & $78.7 \pm 0.1$ & $71.5 \pm 0.2$ & $73.0 \pm 0.5$ & $67.3 \pm 1.1$ & $91.0 \pm 1.5$ & $83.7 \pm 3.8$ & $89.5 \pm 1.3$ & $83.0 \pm 1.4$ \\
\textbf{PGD} & \textbf{BIM} & $80.8 \pm 0.7$ & $73.2 \pm 0.0$ & $78.2 \pm 3.4$ & $70.7 \pm 0.0$ & $72.4 \pm 1.8$ & $66.9 \pm 1.8$ & $90.4 \pm 2.2$ & $83.4 \pm 1.9$ & $89.5 \pm 1.2$ & $82.0 \pm 1.4$ \\
\textbf{PGD} & \textbf{AA} & $81.5 \pm 1.0$ & $65.2 \pm 2.1$ & $78.4 \pm 2.9$ & $63.4 \pm 9.7$ & $84.9 \pm 0.9$ & $68.6 \pm 2.6$ & $61.6 \pm 0.4$ & $54.7 \pm 0.5$ & $57.5 \pm 1.2$ & $52.3 \pm 0.3$ \\
\textbf{PGD} & \textbf{DF} & $77.5 \pm 0.6$ & $62.9 \pm 4.1$ & $98.8 \pm 0.0$ & $94.1 \pm 0.4$ & $99.9 \pm 0.0$ & $95.8 \pm 0.5$ & $43.6 \pm 0.5$ & $48.2 \pm 0.0$ & $45.2 \pm 0.5$ & $48.2 \pm 1.0$ \\
\textbf{PGD} & \textbf{CW} & $72.6 \pm 5.6$ & $57.6 \pm 1.0$ & $97.8 \pm 0.1$ & $92.3 \pm 0.0$ & $99.9 \pm 0.0$ & $97.3 \pm 0.3$ & $42.3 \pm 0.8$ & $48.4 \pm 0.2$ & $44.7 \pm 0.9$ & $48.3 \pm 0.1$ \\ 
\hline 
\textbf{PGD} & \textbf{$\mu$} & $82.8 \pm 0.7$ & $75.8 \pm 1.5$ & $82.2 \pm 1.9$ & $75.2 \pm 1.0$ & $72.8 \pm 1.3$ & $60.8 \pm 3.0$ & $73.0 \pm 0.3$ & $69.8 \pm 1.2$ & $71.4 \pm 1.5$ & $68.8 \pm 0.3$\\ 
\hline

\textbf{AA} & \textbf{FGSM} & $68.1 \pm 0.9$ & $52.4 \pm 1.0$ & $97.4 \pm 0.2$ & $91.6 \pm 0.1$ & $98.5 \pm 0.0$ & $94.8 \pm 0.7$ & $41.6 \pm 0.9$ & $48.9 \pm 0.1$ & $43.8 \pm 0.7$ & $48.3 \pm 0.6$ \\
\textbf{AA} & \textbf{BIM} & $86.8 \pm 2.6$ & $79.6 \pm 3.7$ & $43.8 \pm 12.1$ & $39.2 \pm 1.4$ & $37.8 \pm 0.1$ & $34.3 \pm 1.2$ & $35.7 \pm 0.6$ & $32.9 \pm 0.5$ & $69.4 \pm 1.3$ & $63.0 \pm 0.5$ \\
\textbf{AA} & \textbf{PGD} & $83.3 \pm 4.3$ & $75.1 \pm 0.9$ & $38.7 \pm 3.0$ & $36.5 \pm 0.2$ & $33.6 \pm 5.1$ & $32.6 \pm 1.4$ & $30.6 \pm 7.4$ & $30.0 \pm 1.1$ & $75.2 \pm 1.1$ & $68.0 \pm 0.3$ \\
\textbf{AA} & \textbf{DF} & $77.4 \pm 2.8$ & $70.1 \pm 3.6$ & $80.1 \pm 3.7$ & $72.5 \pm 3.1$ & $81.0 \pm 3.9$ & $73.1 \pm 4.9$ & $64.6 \pm 0.8$ & $57.9 \pm 1.8$ & $63.6 \pm 0.3$ & $56.1 \pm 1.6$ \\
\textbf{AA} & \textbf{CW} & $77.6 \pm 0.9$ & $69.9 \pm 0.0$ & $87.8 \pm 0.3$ & $78.8 \pm 0.3$ & $89.4 \pm 2.2$ & $80.6 \pm 6.8$ & $58.5 \pm 3.9$ & $53.9 \pm 2.5$ & $60.5 \pm 1.1$ & $55.6 \pm 0.9$ \\ 
\hline 
\textbf{AA} & \textbf{$\mu$} & $69.9 \pm 0.5$ & $67.2 \pm 0.5$ & $54.7 \pm 3.3$ & $49.8 \pm 1.5$ & $52.3 \pm 4.2$ & $48.4 \pm 0.8$ & $73.3 \pm 2.3$ & $65.9 \pm 3.0$ & $74.8 \pm 1.7$ & $67.7 \pm 2.1$\\ 
\hline

\textbf{DF} & \textbf{FGSM} & $75.2 \pm 3.3$ & $64.6 \pm 4.9$ & $83.4 \pm 0.7$ & $74.7 \pm 0.1$ & $88.0 \pm 1.0$ & $79.9 \pm 0.4$ & $57.6 \pm 0.3$ & $52.4 \pm 0.2$ & $61.6 \pm 0.1$ & $55.3 \pm 0.3$ \\
\textbf{DF} & \textbf{BIM} & $74.8 \pm 7.1$ & $65.2 \pm 11.4$ & $83.1 \pm 2.6$ & $74.7 \pm 1.6$ & $87.4 \pm 0.4$ & $78.8 \pm 0.5$ & $55.1 \pm 1.4$ & $50.6 \pm 0.4$ & $57.1 \pm 0.4$ & $52.4 \pm 0.1$ \\
\textbf{DF} & \textbf{PGD} & $74.2 \pm 0.3$ & $68.1 \pm 0.1$ & $67.2 \pm 2.8$ & $64.2 \pm 0.2$ & $68.4 \pm 7.4$ & $64.5 \pm 7.3$ & $65.5 \pm 0.8$ & $63.4 \pm 0.3$ & $67.9 \pm 0.7$ & $61.9 \pm 0.7$ \\
\textbf{DF} & \textbf{AA} & $73.5 \pm 0.7$ & $66.7 \pm 0.3$ & $72.8 \pm 0.7$ & $67.7 \pm 0.9$ & $78.0 \pm 1.7$ & $72.2 \pm 3.2$ & $71.4 \pm 5.6$ & $65.1 \pm 2.5$ & $68.7 \pm 1.4$ & $62.9 \pm 5.4$ \\
\textbf{DF} & \textbf{CW} & $70.4 \pm 0.0$ & $63.8 \pm 0.6$ & $69.2 \pm 5.0$ & $63.4 \pm 14.9$ & $75.8 \pm 17.0$ & $69.2 \pm 19.4$ & $51.6 \pm 0.3$ & $51.2 \pm 0.0$ & $52.0 \pm 0.0$ & $51.4 \pm 0.0$ \\ 
\hline 
\textbf{DF} & \textbf{$\mu$} & $73.2 \pm 1.1$ & $65.4 \pm 1.2$ & $71.5 \pm 2.4$ & $64.3 \pm 2.8$ & $68.6 \pm 2.4$ & $64.4 \pm 1.7$ & $72.9 \pm 2.0$ & $66.9 \pm 2.5$ & $63.8 \pm 4.5$ & $59.8 \pm 7.0$\\ 
\hline

\textbf{CW} & \textbf{FGSM} & $59.8 \pm 1.0$ & $54.5 \pm 2.1$ & $93.1 \pm 1.5$ & $85.4 \pm 5.1$ & $99.8 \pm 0.0$ & $92.1 \pm 7.5$ & $54.4 \pm 0.4$ & $52.4 \pm 0.1$ & $54.2 \pm 0.4$ & $51.7 \pm 0.2$ \\
\textbf{CW} & \textbf{BIM} & $58.7 \pm 0.6$ & $53.5 \pm 0.4$ & $92.6 \pm 3.6$ & $84.3 \pm 5.0$ & $99.8 \pm 0.0$ & $93.4 \pm 0.2$ & $54.3 \pm 0.1$ & $51.9 \pm 0.1$ & $54.0 \pm 0.6$ & $51.6 \pm 0.1$ \\
\textbf{CW} & \textbf{PGD} & $55.8 \pm 0.5$ & $50.2 \pm 0.1$ & $91.6 \pm 0.2$ & $75.7 \pm 5.7$ & $93.3 \pm 0.1$ & $77.3 \pm 1.6$ & $53.8 \pm 0.1$ & $50.7 \pm 0.0$ & $53.7 \pm 0.6$ & $50.4 \pm 0.0$ \\
\textbf{CW} & \textbf{AA} & $57.5 \pm 2.7$ & $55.8 \pm 1.5$ & $88.1 \pm 3.4$ & $72.2 \pm 3.2$ & $85.4 \pm 9.9$ & $73.0 \pm 0.0$ & $97.0 \pm 0.6$ & $75.1 \pm 10.1$ & $53.6 \pm 0.2$ & $52.2 \pm 0.2$ \\
\textbf{CW} & \textbf{DF} & $58.6 \pm 4.9$ & $56.3 \pm 3.5$ & $87.4 \pm 1.7$ & $71.5 \pm 5.0$ & $83.4 \pm 9.5$ & $72.0 \pm 8.6$ & $96.1 \pm 2.4$ & $75.8 \pm 1.5$ & $53.6 \pm 0.0$ & $52.6 \pm 1.3$ \\ 
\hline 
\textbf{CW} & \textbf{$\mu$} & $72.2 \pm 0.7$ & $67.2 \pm 3.0$ & $71.9 \pm 1.0$ & $66.9 \pm 1.2$ & $69.6 \pm 0.3$ & $60.8 \pm 1.5$ & $76.3 \pm 3.4$ & $65.6 \pm 3.0$ & $75.8 \pm 3.7$ & $65.7 \pm 4.0$\\ 
\hline 
\multicolumn{12}{c}{\textbf{random forest}} \\ \hline 
\textbf{FGSM} & \textbf{BIM} & $62.2 \pm 249.6$ & $56.4 \pm 131.6$ & $86.5 \pm 1.8$ & $70.8 \pm 23.6$ & $82.2 \pm 336.8$ & $76.6 \pm 257.7$ & $84.4 \pm 2.0$ & $72.6 \pm 1.4$ & $81.2 \pm 1.3$ & $68.5 \pm 4.4$ \\
\textbf{FGSM} & \textbf{PGD} & $60.6 \pm 321.0$ & $56.3 \pm 154.8$ & $68.8 \pm 389.8$ & $66.2 \pm 317.8$ & $83.3 \pm 347.0$ & $80.7 \pm 316.8$ & $63.6 \pm 124.9$ & $54.7 \pm 2.2$ & $58.6 \pm 130.2$ & $52.2 \pm 3.7$ \\
\textbf{FGSM} & \textbf{AA} & $84.8 \pm 0.1$ & $73.0 \pm 3.5$ & $69.5 \pm 301.0$ & $65.2 \pm 250.2$ & $85.5 \pm 362.8$ & $83.3 \pm 324.8$ & $84.2 \pm 1.5$ & $57.6 \pm 3.5$ & $84.8 \pm 0.4$ & $57.2 \pm 0.9$ \\
\textbf{FGSM} & \textbf{DF} & $78.0 \pm 263.0$ & $69.2 \pm 100.4$ & $77.7 \pm 116.9$ & $61.2 \pm 2.4$ & $80.1 \pm 289.1$ & $68.8 \pm 77.4$ & $80.5 \pm 288.1$ & $62.4 \pm 44.3$ & $78.7 \pm 385.7$ & $60.5 \pm 49.6$ \\
\textbf{FGSM} & \textbf{CW} & $86.6 \pm 1.2$ & $78.7 \pm 1.7$ & $61.3 \pm 190.5$ & $49.7 \pm 0.1$ & $84.0 \pm 5.4$ & $55.3 \pm 0.3$ & $79.9 \pm 397.6$ & $74.5 \pm 299.3$ & $97.7 \pm 0.0$ & $95.4 \pm 0.1$ \\ 
\hline 
\textbf{FGSM} & \textbf{$\mu$} & $79.3 \pm 118.3$ & $69.0 \pm 83.8$ & $67.0 \pm 262.6$ & $62.0 \pm 159.1$ & $81.8 \pm 133.2$ & $67.3 \pm 116.6$ & $79.0 \pm 268.5$ & $64.4 \pm 54.8$ & $81.9 \pm 118.9$ & $70.7 \pm 60.3$\\ 
\hline

\textbf{BIM} & \textbf{FGSM} & $84.5 \pm 3.5$ & $77.9 \pm 1.2$ & $49.7 \pm 186.9$ & $49.0 \pm 0.2$ & $78.5 \pm 2.1$ & $53.8 \pm 3.1$ & $76.3 \pm 567.8$ & $69.5 \pm 221.4$ & $97.2 \pm 0.1$ & $94.5 \pm 0.2$ \\
\textbf{BIM} & \textbf{PGD} & $77.6 \pm 0.0$ & $67.5 \pm 0.1$ & $74.9 \pm 1.4$ & $65.7 \pm 0.9$ & $88.8 \pm 0.5$ & $81.0 \pm 1.4$ & $80.9 \pm 1.6$ & $72.2 \pm 0.2$ & $81.4 \pm 1.6$ & $69.5 \pm 0.9$ \\
\textbf{BIM} & \textbf{AA} & $69.7 \pm 2.8$ & $57.1 \pm 1.3$ & $91.1 \pm 2.0$ & $83.4 \pm 4.8$ & $97.4 \pm 0.8$ & $92.3 \pm 1.8$ & $64.3 \pm 12.2$ & $52.8 \pm 0.3$ & $64.7 \pm 2.4$ & $52.4 \pm 0.9$ \\
\textbf{BIM} & \textbf{DF} & $67.5 \pm 2.9$ & $57.1 \pm 2.3$ & $93.5 \pm 0.4$ & $86.2 \pm 0.4$ & $97.0 \pm 0.9$ & $91.3 \pm 1.5$ & $58.3 \pm 7.0$ & $51.8 \pm 0.2$ & $58.8 \pm 4.8$ & $51.9 \pm 2.3$ \\
\textbf{BIM} & \textbf{CW} & $79.7 \pm 0.4$ & $57.4 \pm 2.4$ & $91.8 \pm 0.8$ & $83.7 \pm 1.2$ & $88.3 \pm 4.6$ & $79.4 \pm 4.1$ & $79.9 \pm 0.9$ & $54.4 \pm 0.6$ & $79.6 \pm 1.7$ & $54.0 \pm 0.6$ \\ 
\hline 
\textbf{BIM} & \textbf{$\mu$} & $77.2 \pm 152.1$ & $69.0 \pm 45.2$ & $80.7 \pm 1.0$ & $71.2 \pm 0.7$ & $77.4 \pm 4.1$ & $67.6 \pm 1.8$ & $75.0 \pm 3.2$ & $67.7 \pm 1.4$ & $83.9 \pm 1.7$ & $65.8 \pm 1.8$\\ 
\hline

\textbf{PGD} & \textbf{FGSM} & $81.2 \pm 0.2$ & $72.7 \pm 0.0$ & $71.6 \pm 1.6$ & $63.7 \pm 3.0$ & $66.7 \pm 1.6$ & $57.6 \pm 2.6$ & $82.3 \pm 4.5$ & $75.8 \pm 25.4$ & $89.7 \pm 0.5$ & $83.1 \pm 2.5$ \\
\textbf{PGD} & \textbf{BIM} & $79.8 \pm 0.6$ & $72.1 \pm 0.7$ & $69.8 \pm 0.0$ & $58.1 \pm 9.1$ & $64.9 \pm 2.8$ & $53.2 \pm 5.4$ & $79.0 \pm 13.8$ & $67.2 \pm 47.7$ & $89.1 \pm 1.4$ & $82.8 \pm 1.9$ \\
\textbf{PGD} & \textbf{AA} & $80.1 \pm 1.1$ & $65.1 \pm 1.5$ & $76.6 \pm 0.5$ & $64.1 \pm 0.7$ & $84.3 \pm 0.1$ & $70.1 \pm 1.5$ & $54.9 \pm 0.7$ & $52.3 \pm 0.0$ & $52.8 \pm 0.2$ & $51.3 \pm 0.2$ \\
\textbf{PGD} & \textbf{DF} & $81.6 \pm 6.1$ & $56.8 \pm 7.9$ & $98.8 \pm 0.1$ & $95.0 \pm 0.8$ & $99.9 \pm 0.0$ & $97.0 \pm 0.1$ & $44.2 \pm 4.8$ & $48.2 \pm 0.1$ & $43.3 \pm 1.0$ & $47.8 \pm 0.8$ \\
\textbf{PGD} & \textbf{CW} & $83.0 \pm 0.2$ & $52.6 \pm 1.3$ & $97.4 \pm 0.2$ & $90.8 \pm 0.5$ & $99.9 \pm 0.0$ & $97.9 \pm 0.0$ & $44.6 \pm 4.2$ & $48.6 \pm 0.0$ & $44.2 \pm 0.9$ & $48.5 \pm 0.5$ \\ 
\hline 
\textbf{PGD} & \textbf{$\mu$} & $78.3 \pm 1.7$ & $70.6 \pm 6.7$ & $76.5 \pm 3.7$ & $66.7 \pm 13.0$ & $69.7 \pm 0.6$ & $60.6 \pm 0.8$ & $73.6 \pm 2.4$ & $69.0 \pm 1.9$ & $73.8 \pm 1.1$ & $67.7 \pm 0.5$\\ 
\hline

\textbf{AA} & \textbf{FGSM} & $76.4 \pm 4.0$ & $50.3 \pm 0.0$ & $97.0 \pm 0.3$ & $88.3 \pm 0.0$ & $98.5 \pm 0.1$ & $92.2 \pm 0.8$ & $45.8 \pm 2.4$ & $49.5 \pm 0.0$ & $44.9 \pm 6.1$ & $49.2 \pm 0.1$ \\
\textbf{AA} & \textbf{BIM} & $76.3 \pm 3.3$ & $69.5 \pm 8.4$ & $19.3 \pm 21.3$ & $34.4 \pm 1.1$ & $15.5 \pm 11.8$ & $32.5 \pm 0.3$ & $13.6 \pm 12.8$ & $31.7 \pm 0.9$ & $69.3 \pm 0.5$ & $63.7 \pm 0.6$ \\
\textbf{AA} & \textbf{PGD} & $70.7 \pm 4.7$ & $66.6 \pm 7.5$ & $21.0 \pm 5.0$ & $32.7 \pm 2.2$ & $17.8 \pm 12.6$ & $30.1 \pm 2.5$ & $15.0 \pm 4.2$ & $29.6 \pm 3.8$ & $73.1 \pm 0.7$ & $67.7 \pm 1.1$ \\
\textbf{AA} & \textbf{DF} & $76.0 \pm 3.3$ & $69.6 \pm 1.0$ & $78.5 \pm 2.8$ & $71.0 \pm 3.0$ & $78.1 \pm 2.8$ & $70.2 \pm 2.8$ & $61.1 \pm 1.2$ & $57.6 \pm 1.1$ & $60.7 \pm 0.4$ & $57.2 \pm 2.7$ \\
\textbf{AA} & \textbf{CW} & $71.6 \pm 2.4$ & $63.9 \pm 1.3$ & $86.1 \pm 1.2$ & $79.1 \pm 0.8$ & $89.2 \pm 1.0$ & $81.0 \pm 1.7$ & $53.9 \pm 0.3$ & $51.5 \pm 1.0$ & $58.2 \pm 2.0$ & $54.7 \pm 0.6$ \\ 
\hline 
\textbf{AA} & \textbf{$\mu$} & $72.5 \pm 2.6$ & $65.9 \pm 0.2$ & $38.8 \pm 9.9$ & $46.4 \pm 2.3$ & $39.5 \pm 5.5$ & $45.3 \pm 3.4$ & $70.9 \pm 2.1$ & $65.1 \pm 2.1$ & $71.8 \pm 1.4$ & $66.0 \pm 1.1$\\ 
\hline

\textbf{DF} & \textbf{FGSM} & $70.0 \pm 1.1$ & $60.3 \pm 2.9$ & $80.5 \pm 2.1$ & $71.9 \pm 2.4$ & $87.5 \pm 0.9$ & $78.8 \pm 2.6$ & $54.6 \pm 0.6$ & $51.4 \pm 0.3$ & $58.8 \pm 0.6$ & $54.5 \pm 0.7$ \\
\textbf{DF} & \textbf{BIM} & $65.9 \pm 0.7$ & $60.0 \pm 3.3$ & $81.4 \pm 2.7$ & $72.8 \pm 5.5$ & $86.7 \pm 0.6$ & $78.3 \pm 1.0$ & $52.4 \pm 0.6$ & $50.5 \pm 0.2$ & $56.5 \pm 1.1$ & $53.1 \pm 0.0$ \\
\textbf{DF} & \textbf{PGD} & $66.0 \pm 1.5$ & $61.3 \pm 1.0$ & $58.5 \pm 2.9$ & $53.9 \pm 6.8$ & $63.5 \pm 4.1$ & $57.1 \pm 2.8$ & $55.0 \pm 2.6$ & $50.0 \pm 1.5$ & $75.0 \pm 0.4$ & $69.1 \pm 0.3$ \\
\textbf{DF} & \textbf{AA} & $66.6 \pm 7.5$ & $63.1 \pm 1.9$ & $65.2 \pm 7.3$ & $61.2 \pm 9.4$ & $70.3 \pm 2.7$ & $66.1 \pm 3.3$ & $62.2 \pm 10.5$ & $59.0 \pm 13.3$ & $70.7 \pm 0.5$ & $65.4 \pm 0.6$ \\
\textbf{DF} & \textbf{CW} & $55.3 \pm 3.7$ & $52.8 \pm 3.6$ & $59.5 \pm 1.3$ & $55.8 \pm 1.9$ & $50.7 \pm 11.7$ & $48.8 \pm 5.8$ & $50.1 \pm 0.3$ & $49.6 \pm 0.1$ & $50.5 \pm 0.2$ & $50.6 \pm 0.0$ \\ 
\hline 
\textbf{DF} & \textbf{$\mu$} & $70.3 \pm 1.1$ & $63.4 \pm 1.8$ & $68.6 \pm 1.1$ & $62.9 \pm 2.0$ & $63.6 \pm 2.3$ & $58.3 \pm 2.5$ & $67.0 \pm 5.7$ & $63.0 \pm 5.7$ & $53.2 \pm 3.4$ & $51.5 \pm 2.3$\\ 
\hline

\textbf{CW} & \textbf{FGSM} & $54.6 \pm 0.6$ & $51.6 \pm 0.0$ & $94.1 \pm 0.1$ & $87.3 \pm 1.8$ & $99.5 \pm 0.0$ & $93.4 \pm 1.6$ & $53.9 \pm 0.1$ & $52.5 \pm 0.0$ & $53.4 \pm 0.5$ & $51.5 \pm 0.2$ \\
\textbf{CW} & \textbf{BIM} & $55.2 \pm 1.0$ & $51.5 \pm 0.3$ & $91.8 \pm 0.6$ & $82.9 \pm 1.3$ & $99.6 \pm 0.0$ & $94.2 \pm 1.6$ & $53.6 \pm 0.3$ & $51.7 \pm 0.4$ & $53.7 \pm 0.4$ & $51.2 \pm 0.2$ \\
\textbf{CW} & \textbf{PGD} & $51.4 \pm 2.6$ & $49.9 \pm 0.1$ & $85.7 \pm 0.7$ & $63.4 \pm 5.8$ & $89.9 \pm 0.4$ & $71.9 \pm 0.1$ & $51.5 \pm 1.1$ & $50.2 \pm 0.0$ & $51.7 \pm 0.1$ & $50.3 \pm 0.0$ \\
\textbf{CW} & \textbf{AA} & $52.8 \pm 5.4$ & $51.5 \pm 0.5$ & $74.3 \pm 2.6$ & $67.1 \pm 2.1$ & $78.6 \pm 15.6$ & $69.5 \pm 6.9$ & $75.1 \pm 227.1$ & $66.4 \pm 84.9$ & $53.6 \pm 0.2$ & $52.0 \pm 0.9$ \\
\textbf{CW} & \textbf{DF} & $53.4 \pm 1.1$ & $53.0 \pm 1.8$ & $76.8 \pm 7.9$ & $67.7 \pm 0.8$ & $80.8 \pm 21.8$ & $71.2 \pm 15.1$ & $87.5 \pm 63.7$ & $72.5 \pm 5.4$ & $50.9 \pm 0.8$ & $50.8 \pm 1.5$ \\ 
\hline 
\textbf{CW} & \textbf{$\mu$} & $71.1 \pm 0.3$ & $67.3 \pm 0.7$ & $70.8 \pm 0.5$ & $66.3 \pm 0.8$ & $66.0 \pm 1.0$ & $57.1 \pm 1.2$ & $66.9 \pm 50.2$ & $61.3 \pm 19.1$ & $69.9 \pm 19.0$ & $63.0 \pm 4.9$\\ 
\hline
\bottomrule[1pt]
\end{tabular}
}
\end{adjustbox}
\end{table*}

\end{document}